\documentclass{article} 

\usepackage{iclr2027_conference,times}
\usepackage{graphicx}

\usepackage{amsmath,amssymb}

\usepackage{booktabs}
\usepackage{multirow}

\usepackage[ruled,vlined,linesnumbered]{algorithm2e}
\usepackage{float}

\usepackage{wrapfig}

\usepackage[table]{xcolor}
\usepackage{enumitem}
\usepackage{tcolorbox}
\tcbuselibrary{skins,breakable}
\usepackage{pgfplots}
\pgfplotsset{compat=1.18}

\usepackage{xurl}
\usepackage{titletoc}
\usepackage{hyperref}
\usepackage[all]{hypcap}

\definecolor{fullcolor}{HTML}{3B4252}
\definecolor{nocolor}{HTML}{EBCB8B}
\definecolor{uniformcolor}{HTML}{A3BE8C}
\definecolor{madcolor}{HTML}{BF616A}
\definecolor{referencecolor}{HTML}{D8DEE9}

\SetAlgoNlRelativeSize{-2}
\SetNlSkip{0.25em}
\SetKwInOut{KwIn}{Input}
\SetKwInOut{KwOut}{Output}

\setcitestyle{yysep={,}}

\newtcolorbox{lacpdpromptbox}[1]{
  enhanced,
  breakable,
  sharp corners,
  colback=black!2,
  colframe=black,
  colbacktitle=black,
  coltitle=white,
  boxrule=0.5pt,
  titlerule=0pt,
  fonttitle=\bfseries\small,
  fontupper=\small,
  title={#1},
  title after break={#1 (continued)},
  left=8pt,
  right=8pt,
  top=7pt,
  bottom=7pt,
  toptitle=3pt,
  bottomtitle=3pt,
  before skip=10pt,
  after skip=10pt,
  parbox=false,
  before upper={
    \setlength{\parindent}{0pt}
    \setlength{\parskip}{3pt}
    \raggedright
  }
}

\newcommand{\lacpdpromptheading}[1]{%
  \par\noindent\textbf{#1}\par\nopagebreak
}

\newcommand{\lacpdpromptdivider}{%
  \par\medskip
  {\color{black!55}\hrule height 0.4pt}
  \medskip
}

\newlist{lacpdpromptitems}{itemize}{1}
\setlist[lacpdpromptitems]{
  label={\textnormal{-}},
  leftmargin=1em,
  labelsep=0.35em,
  itemsep=0pt,
  parsep=0pt,
  topsep=2pt,
  partopsep=0pt
}

\usepackage{amsmath,amsfonts,bm}

\def\eqref#1{equation~\ref{#1}}

\def\1{\bm{1}}

\DeclareMathAlphabet{\mathsfit}{\encodingdefault}{\sfdefault}{m}{sl}
\SetMathAlphabet{\mathsfit}{bold}{\encodingdefault}{\sfdefault}{bx}{n}

\title{LA-CPD: Local-Evidence-Aware Change-Point Detection for Human–LLM Authorship Segmentation}

\author{
Qing~Yang$^{1}$, Zhenyu~Mao$^{1}$, 
Zixiang~Luo$^{1}$, Zezheng~Wu$^{1}$,
Xinghe~Cheng$^{2}$,Qinggang Zhang$^{3}$ \\
\textbf{Jingwei Zhang}$^{1}$\thanks{Corresponding author.},
\textbf{Jiapu Wang}$^{4}$\\
$^{1}$Guilin University of Electronic Technology, China,\qquad
$^{2}$Jinan Unaiversity, China,\\
$^{3}$Jilin University, China,\qquad
$^{4}$Nanjing University of Science and Technology, China\\
\texttt{\{gtyqing,gtzjw\}@hotmail.com},\qquad
\texttt{mzy@mails.guet.edu.cn},\\
\texttt{jnuchengxh@hotmail.com},\qquad
\texttt{jiapu.wang@njust.edu.cn}
}

\iclrfinalcopy

\begin{document}

\maketitle

\lhead{Preprint}

\begin{abstract}



As LLM-generated text becomes increasingly human-like, failing to localize LLM-authored spans in human--LLM co-authored documents may weaken our ability to establish accountability for copyright infringement, fraud, and other harmful uses of AI-generated content. Existing sentence-level detection scores provide local evidence for these transitions, but content differences can cause score fluctuations even among sentences from the same source, potentially introducing spurious boundaries. Identifying genuine authorship transitions and delineating the corresponding document partition therefore remain challenging when both boundary locations and counts are unknown.
We propose \textbf{L}ocal-Evidence-\textbf{A}ware \textbf{C}hange-\textbf{P}oint \textbf{D}etection (\textbf{LA-CPD}), a structured method that converts noisy score sequences into coherent authorship segments. Given scores from a frozen local detector, LA-CPD combines a length-weighted within-segment residual with a windowed two-mean contrast to capture both segment consistency and sustained changes across candidate cuts. Dynamic programming optimizes cut locations for each candidate count, while an AIC-style criterion selects the final partition for delineating sentence labels, authorship boundaries, and maximal LLM-authored spans. Experiments on a held-out human–LLM co-authored test set show that LA-CPD outperforms WCP+AIC, improving sentence-level accuracy from $0.747$ to $0.796$ while achieving better boundary localization and LLM-span delineation.\footnote{Code and data are available at:
\href{https://anonymous.4open.science/r/anonymous-submission-D407/}
{https://anonymous.4open.science}.}

\end{abstract}

\section{Introduction}
\label{sec:introduction}
Large Language Models (LLMs) ~\citep{pan2024unifying,zhao2026survey} are increasingly used to generate or revise selected portions of human-authored documents, allowing human and LLM contributions to coexist within a single text ~\citep{cheng2025beyondbinary,su2025hacodet}. These contributions may alternate several times throughout a document, yielding multiple authorship transitions ~\citep{zeng2024detecting,zeng2024boundary}. This internal provenance structure cannot be represented by a document-level label, which can indicate machine involvement but not the location or extent of LLM-generated content ~\citep{lei2025pald}. Sentence-level authorship segmentation is therefore needed to delineate contiguous human- and LLM-authored segments and their transition boundaries.

Sentence-level detection scores provide a natural source of evidence for delineating this internal structure ~\citep{wang2023seqxgpt,su2025hacodet}.
However, these scores are not determined by source identity alone.
Their values vary with sentence-specific content, while short sentences may provide less stable evidence ~\citep{fraser2025detecting}.
Consequently, substantial score fluctuations can occur even across consecutive sentences written by the same source.
Methods that infer boundaries from individual scores or adjacent differences may mistake these fluctuations for authorship changes, producing spurious boundaries and fragmenting contiguous LLM-authored spans ~\citep{wang2023seqxgpt,jiang2025sendetex,li2026segmenting}.

Reliable segmentation must therefore reason over regions rather than isolated sentence positions ~\citep{truong2020selective}.
A credible boundary should separate regions that are internally consistent yet differ persistently in score level.
These requirements are complementary: optimizing only within-segment fit favors increasingly fine partitions, whereas adjacent-score contrast alone is sensitive to isolated deviations ~\citep{li2026segmenting}.
Moreover, neither the locations nor the number of boundaries are known in advance ~\citep{yao1988estimating,jackson2005algorithm}.
Authorship localization therefore requires selecting a complete partition whose improvement in fit justifies its complexity.

To address these challenges, we propose \textbf{Local-Evidence-Aware Change-Point Detection (LA-CPD)}, a structured authorship segmentation method for documents with unknown boundary locations and counts~\citep{truong2020selective}.
LA-CPD preserves document order and represents the sentence scores produced by a frozen local detector as an authorship-evidence trajectory.
It evaluates candidate partitions using a length-weighted within-segment residual and a windowed two-mean contrast.
The former measures whether the scores within a segment are compatible with a shared evidence level, whereas the latter measures whether a candidate cut is supported by a sustained difference across its neighboring sentences.
Their combination allows each cut to be assessed through both the consistency of the resulting segments and the persistence of the corresponding transition.

For each candidate cut count, dynamic programming identifies the globally optimal cut locations under the proposed objective.
An AIC-style criterion then selects among the resulting partitions by balancing residual fit against model complexity ~\citep{jackson2005algorithm,yao1988estimating}.
Finally, LA-CPD assigns Human/LLM labels from the mean scores of the selected segments and merges adjacent segments with the same label, yielding the final authorship boundaries and maximal contiguous LLM-authored spans.
To isolate the contribution of structured segmentation, all score-based methods in our experiments use the same frozen detector and identical score trajectories.

The main contributions of this work are summarized as follows:

\begin{itemize}[
    leftmargin=1.2em,
    labelindent=0pt,
    labelwidth=0.7em,
    labelsep=0.5em,
    itemsep=2pt,
    topsep=2pt
]
    \item To the best of our knowledge, LA-CPD is the first human--LLM authorship segmentation method to incorporate fixed-window transition evidence directly into a globally optimized partition objective;

    \item LA-CPD combines the length-weighted within-segment residual $V(s,e)$ with the windowed two-mean contrast $E(c;h)$, evaluating candidate cuts through both segment consistency and sustained cross-boundary change;

    \item LA-CPD develops a two-stage inference procedure for unknown boundary counts: dynamic programming obtains the optimal partition for each candidate count, and an AIC-style criterion selects among these candidates before authorship-label and LLM-span delineation.
\end{itemize}
We develop an end-to-end inference procedure for unknown boundary
counts: dynamic programming optimizes cut locations for each candidate
number of statistical cuts, an AIC-style criterion selects the final
partition, and segment-level clustering delineates source labels and
authorship spans.

\section{Related Work}
\label{sec:related-work}


\paragraph{LLM-Generated Text Detection.}
Existing detectors derive signals from token probabilities or ranks,
probability curvature, regeneration discrepancies, and
representation-level statistics
~\citep{gehrmann2019gltr,mitchell2023detectgpt,bao2024fast,tulchinskii2023intrinsic,xu2025tokenprob,zhou2025adadetectgpt,nassif2026telescope,wu2026alignmentimprint}. For machine-revised and human--machine hybrid text, ImBD imitates the stylistic preferences of machine-revised text, whereas MSPO models style differences at multiple linguistic levels
~\citep{chen2025imbd,wang2026mspo}. Many of these methods produce document-level decisions rather than explicit authorship boundaries.

\paragraph{Localization of Human--LLM Co-Authored Text.}
Recent work extends document-level detection to characterize the mode,
extent, and location of machine involvement. Beyond Binary, Beyond the
Final Actor, and DETree study human--LLM collaboration from the
perspectives of authorship roles, creator--editor relations, and
hierarchical representations, respectively ~\citep{cheng2025beyondbinary,li2026beyondfinal,he2025detree}.
PaLD localizes machine-written portions and estimates their proportion,
whereas EditLens estimates the extent of AI editing
~\citep{lei2025pald,thai2026editlens}.

Another line of work studies authorship boundaries in mixed text. RoFT
introduced a human-written-prefix and machine-generated-continuation setting for evaluating human recognition of authorship transitions
~\citep{dugan2020roft}, followed by automatic boundary detection studies
such as ~\citet{kushnareva2024roft}. In educational
writing, ~\citet{zeng2024boundary} study boundary detection
in hybrid essays, while ~\citet{zeng2024detecting} address
sentence-level provenance identification. SeqXGPT, SenDetEX, HACo-Det,
DAMASHA, and GigaCheck further investigate fine-grained localization
using probabilistic, stylistic, attribution-based, and
character-level span representations
~\citep{wang2023seqxgpt,jiang2025sendetex,su2025hacodet,
saiteja2026damasha,tolstykh2026gigacheck}. Change-point-based methods
treat local detection scores as ordered observations; ~\citet{li2026segmenting} formulate human--LLM text localization through VCP, weighted WCP, and GCP.

Although prior work has advanced LLM-text detection, distinguishing genuine authorship transitions from score fluctuations and selecting a complete partition when the number of boundaries is unknown remain challenging. LA-CPD performs two-stage inference for unknown boundary counts: it uses dynamic programming to optimize each candidate partition, and applies an AIC-style criterion to select the final partition before delineating source labels, authorship boundaries, and contiguous LLM spans.

\section{Preliminary}
\label{sec:preliminary}

\textbf{Task Definition} We study sentence-level authorship segmentation for documents with unknown authorship-boundary locations and counts. 

Given a human--LLM co-authored document $\mathcal{D}=(u_1,\ldots,u_n)$, where $u_i$ denotes the $i$-th sentence, each sentence is assigned a
source label $y_i\in\mathcal{Y}=\{\mathrm{H},\mathrm{L}\}$, where $\mathrm{H}$ denotes human authorship and $\mathrm{L}$ denotes LLM authorship. A change in the source labels of two adjacent sentences defines an authorship boundary. The reference boundary set is
\begin{equation}
C=
\left\{
c\in\{1,\ldots,n-1\}:y_c\neq y_{c+1}
\right\},
\label{eq:reference-authorship-boundaries}
\end{equation}
where boundary \(c\) represents an authorship transition between sentences \(u_c\) and \(u_{c+1}\). The reference boundary count is $K^\star=|C|$. At inference time, the model receives only the document \(\mathcal{D}\) and does not access the reference source labels. It predicts the source-label sequence $ \widehat{\boldsymbol y} = (\widehat y_1,\ldots,\widehat y_n).$
The final predicted authorship-boundary set is obtained from changes in the predicted labels:
$
\widehat C=
\left\{
i\in\{1,\ldots,n-1\}:
\widehat y_i\neq \widehat y_{i+1}
\right\}.
$
The predicted boundary count is
$
\widehat K=|\widehat C|.
$
Maximal contiguous sentence intervals whose predicted labels are
\(\mathrm{L}\) constitute the predicted LLM-authored fragments.

\paragraph{Dynamic programming.}
Dynamic programming is a general optimization strategy for problems
with optimal substructure. The classical formulation of dynamic
programming dates back to ~\cite{bellman1966dynamic}.
It decomposes a global problem into overlapping subproblems, solves
each subproblem once, and combines their solutions to obtain a global
optimum. In LA-CPD, it optimizes the cut locations for each candidate cut count $k$;
the choice of $k$ is handled separately by the AIC-style criterion.

\paragraph{AIC-style model selection.}
The Akaike information criterion (AIC) was introduced to select
statistical models by balancing data fit against model complexity~\citep{akaike1974new}. In LA-CPD, we use an AIC-style criterion to compare candidate partitions with different numbers of statistical cuts. Specifically, the criterion combines the proposed residual term with a user-defined complexity multiplier $r$. Therefore, our selection rule is AIC-inspired but is not the standard AIC; the residual definition and the penalty multiplier are specific to LA-CPD.

\section{Methodology}
\label{sec:method}

\begin{figure}[t]
    \centering
    \includegraphics[width=\linewidth]{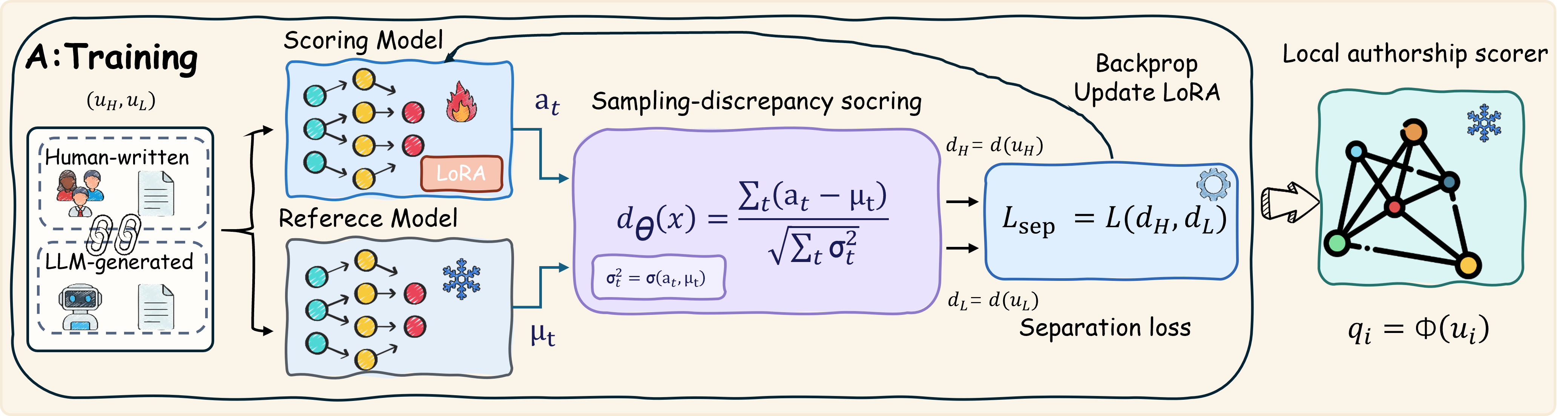}
    \caption{LA-CPD training. Human--LLM sentence pairs are used to adapt a LoRA-augmented scoring model with a frozen reference model through a sampling-discrepancy separation loss. The adapted scorer is then frozen to produce sentence-level scores $q_i=\Phi(u_i)$ for structured segmentation.}
    \label{fig:detector-training}
\end{figure}

\begin{figure}[t]
    \centering
    \includegraphics[width=\linewidth]{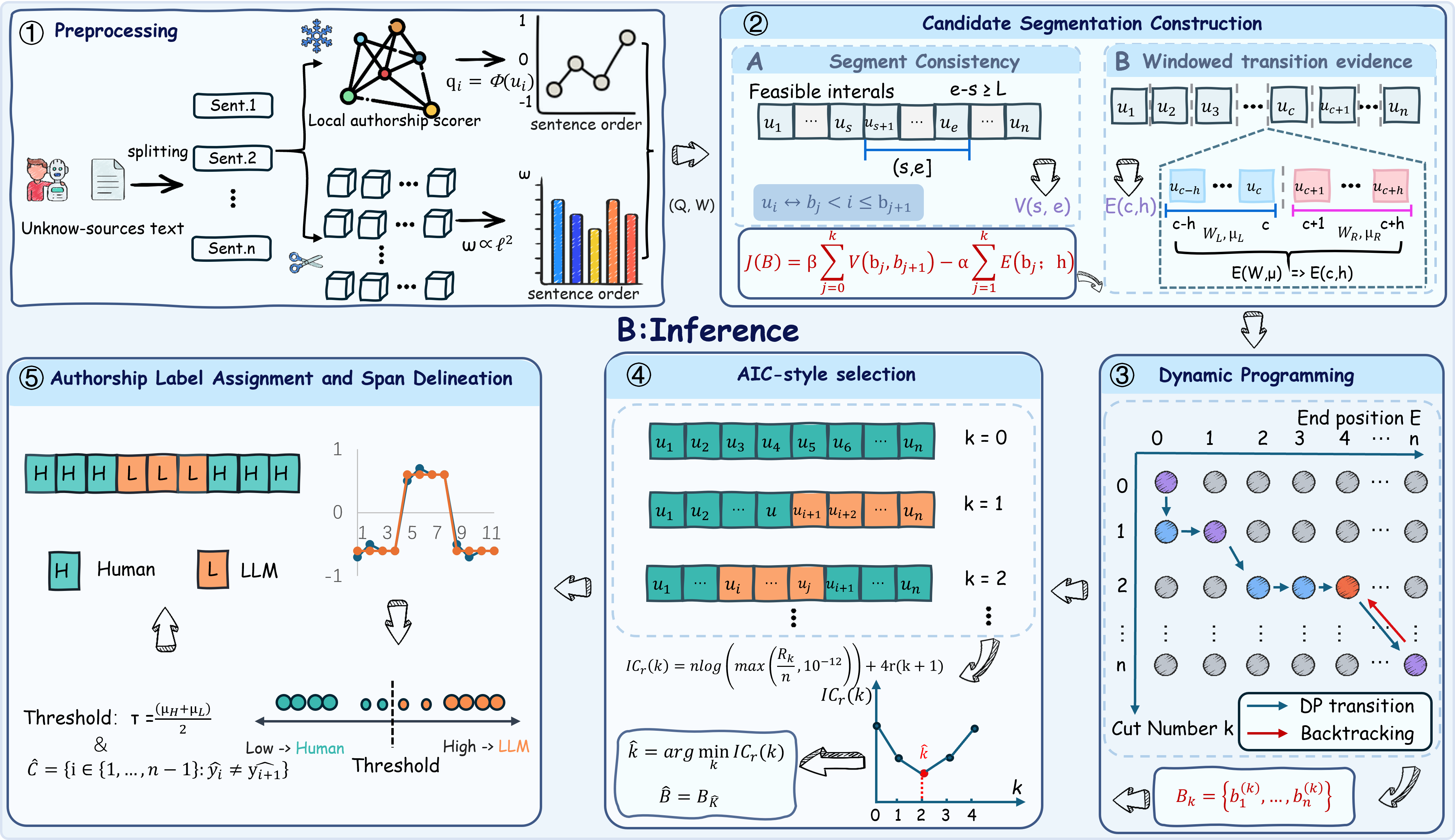}
    \caption{LA-CPD inference. Top left: a frozen local detector produces ordered sentence-level scores, while
sentence lengths determine the fitting weights. Top right: $V(s,e)$ and $E(c;h)$ define the candidate-partition objective. Bottom right: dynamic programming optimizes cut locations for each fixed cut count $k$. Bottom middle: an AIC-style criterion selects the final statistical cut set. Bottom left: two-cluster grouping of segment means delineates Human/LLM labels, authorship boundaries, and contiguous LLM-authored spans.}
    \label{fig:segmentation-inference}
\end{figure}

In this section, we propose \textbf{LA-CPD}, a sentence-level authorship
segmentation method for documents with an unknown number of authorship boundaries. LA-CPD consists of two stages: training and inference. During \textbf{training}, a local authorship detector is adapted with sentence-level supervision to produce sentence-level authorship scores. During \textbf{inference}, these scores are combined with within-segment consistency and windowed transition evidence to construct candidate partitions. An AIC-style criterion then selects the number of statistical cuts, after which sentence-level source labels and contiguous authorship spans are delineated. The training and inference stages are illustrated in Figures~\ref{fig:detector-training} and \ref{fig:segmentation-inference}, respectively.


\subsection{Training}
\label{sec:local-detector}
The local detector maps each sentence to a local authorship score, providing sentence-level evidence for subsequent structured segmentation. We adapt a sampling-discrepancy scoring model with sentence-level authorship supervision and freeze it before segmentation inference. The scoring model uses Gemma-3-1B\footnote{Official model card:
\url{https://huggingface.co/google/gemma-3-1b-pt}.} as its backbone and introduces LoRA adapters into the attention projection layers~\citep{hu2022lora}. An independent copy of the same backbone serves as the reference model and remains frozen throughout training. Human- and LLM-written sentences are used to adapt the scoring model and improve the separability of their local scores.

Let $\{(u_{\mathrm H}^{(b)},u_{\mathrm L}^{(b)})\}_{b=1}^{M}$ denote a minibatch of $M$ human--LLM sentence pairs, where $T_{\mathrm H}^{(b)}$ and $T_{\mathrm L}^{(b)}$ are their respective token lengths. Following the sampling-discrepancy formulation of Fast-DetectGPT~\citep{bao2024fast}, we compute the deviation of the scoring-model log probability $a_{\theta,t}$ from the conditional expectation $\mu_{\theta,t}$ under the reference distribution, aggregate the deviations over each sentence, and normalize them by the square root of the accumulated conditional variance. The mean-score separation
objective can be defined as follows:
\begin{equation}
\label{eq:separation-loss}
\begin{aligned}
\mathcal{L}_{\mathrm{sep}}(\theta)
&=
\frac{1}{M}
\sum_{b=1}^{M}
\bigg[
d_{\theta}\!\left(
\mathcal{M}_{\mathrm{score}}
\!\left(u_{\mathrm H}^{(b)}\right),
\mathcal{M}_{\mathrm{ref}}
\!\left(u_{\mathrm H}^{(b)}\right)
\right)
-
d_{\theta}\!\left(
\mathcal{M}_{\mathrm{score}}
\!\left(u_{\mathrm L}^{(b)}\right),
\mathcal{M}_{\mathrm{ref}}
\!\left(u_{\mathrm L}^{(b)}\right)
\right)
\bigg],
\end{aligned}
\end{equation}
where, $b$ indexes a human--LLM sentence pair in a minibatch of size $M$. The representations $\mathcal{M}_{\mathrm{score}}(u)$ and
$\mathcal{M}_{\mathrm{ref}}(u)$ denote the score-model and
reference-model quantities used to compute the sampling-discrepancy score $d_{\theta}$. The token-level score construction, conditional expectation and variance, and numerical stabilization details are provided in Appendix~\ref{app:score-computation}. Minimizing Eq.~(\ref{eq:separation-loss}) encourages higher scores for LLM-written sentences than for human-written sentences. Only the LoRA parameters of the scoring model are updated; the scoring backbone and the reference model remain frozen. The objective does not directly use document-level boundary annotations.

After adaptation, we freeze the scoring model and denote its scoring function by $q_i=\phi(u_i)$. For a document $\mathcal{D}=(u_1,\ldots,u_n)$, we independently compute $q_i=\phi(u_i)$ for each sentence and retain the document order to form the score trajectory $\boldsymbol{q}=(q_1,\ldots,q_n)$. Higher scores indicate stronger evidence of LLM authorship. The score trajectory is then passed to the structured segmentation module to construct candidate partitions and delineate authorship boundaries and contiguous LLM-authored spans. Complete definitions of the score components, sentence-pair construction, model configuration, and optimization details are provided in Appendices~\ref{app:detector-training}--\ref{app:score-computation}.


\subsection{Inference}
\label{sec:structured-segmentation}

Structured inference converts the sentence-level score trajectory
produced by the frozen detector into contiguous authorship spans when
the locations and number of authorship transitions are unknown.
Specifically, it consists of three components: \textbf{Candidate Segmentation
Construction}, \textbf{Dynamic Programming and AIC-Style Selection}, and
\textbf{Authorship Label Assignment and Span Delineation}. Figure~\ref{fig:segmentation-inference}
illustrates this process.

\paragraph{Candidate Segmentation Construction.}
Candidate partitions are evaluated by how well a common score level
explains each segment and how strongly local evidence supports its cuts.
Given scores $\boldsymbol{q}=(q_1,\ldots,q_n)$, we use sentence token
counts $\ell_i$ to weight segment fitting. For an interval $(s,e]$
containing sentences $s+1,\ldots,e$, let $\bar q_{s:e}$ be its weighted
mean. The segment residual and weights are
\begin{equation}
V(s,e)=\sum_{i=s+1}^{e}w_i(q_i-\bar q_{s:e})^2,
\qquad
w_i=\frac{n\ell_i^2}{\sum_{j=1}^{n}\ell_j^2}.
\label{eq:segment-residual}
\end{equation}
Longer sentences contribute more to fitting, while a smaller residual
indicates greater consistency with a shared score level.


Windowed evidence evaluates local score changes around each potential
cut. Every inter-sentence position $c\in\{1,\ldots,n-1\}$ is a potential
cut, with its selection determined by the partition objective and length
constraints. We take up to $h$ sentences on each side, truncating the
windows at document endpoints. With total weights $W_L,W_R$ and weighted
means $\mu_L,\mu_R$, the evidence is
\begin{equation}
E(c;h)=\frac{W_LW_R}{W_L+W_R}(\mu_L-\mu_R)^2.
\label{eq:windowed-evidence}
\end{equation}
This equals the reduction in weighted residual obtained by fitting
separate means to the two windows instead of a shared mean. For $h=1$,
it reduces to a weighted squared difference between adjacent scores.
For $h>1$, neighborhood averaging aggregates evidence of local level
changes and can reduce sensitivity to isolated score fluctuations. The prefix-statistics implementation and the residual decomposition underlying
the windowed evidence are given in
Appendix~\ref{app:interval-statistics}.


The partition objective combines segment residuals with rewards for
cuts supported by windowed evidence. For ordered cuts
$B=\{b_1,\ldots,b_k\}$, with $b_0=0$ and $b_{k+1}=n$, define
\begin{equation}
J(B)=\beta\sum_{j=0}^{k}V(b_j,b_{j+1})
-\alpha\sum_{j=1}^{k}E(b_j;h),
\label{eq:candidate-objective}
\end{equation}
where $\alpha\geq0$ weights the reward for windowed transition evidence, and $\beta>0$ weights the within-segment residual. Every segment must contain at least $L$ sentences;
when $\alpha>0$, we require $L\geq h$.


\paragraph{Dynamic Programming and AIC-Style Selection.}
Dynamic programming first finds the best cut locations for each fixed
count. Let $F_k(e)$ be the minimum value of $J$ for the first $e$
sentences with exactly $k$ statistical cuts. Initialize
$F_0(e)=\beta V(0,e)$ for $e\geq L$ and assign $+\infty$ to infeasible
states. For $k\geq1$ and $e\geq(k+1)L$,
\begin{equation}
F_k(e)=
\min_{\substack{t\in\mathbb{Z}\\kL\leq t\leq e-L}}
\left\{F_{k-1}(t)+\beta V(t,e)-\alpha E(t;h)\right\}.
\label{eq:dp}
\end{equation}
Fixing the last cut $t$ determines the final segment cost and boundary
reward, leaving an optimal prefix problem. A shared dynamic program
therefore finds a global minimum of $J$ at each feasible fixed $k$.
Backtracking the stored predecessors delineates its candidate partition
$B_k$.


An AIC-style criterion selects among these candidates using their
original weighted residuals. Since $J$ includes boundary rewards, we
recompute the residual of each $B_k$ as $R_k=\sum_{j=0}^{k}V\!\left(b_j^{(k)},b_{j+1}^{(k)}\right), b_0^{(k)}=0, b_{k+1}^{(k)}=n$, and evaluate
\begin{equation}
\mathrm{IC}_r(k)
=n\log\!\left(\max\!\left\{\frac{R_k}{n},10^{-12}\right\}\right)
+4r(k+1).
\label{eq:aic-criterion}
\end{equation}
The effective search range, fixed-count optimality argument,
backtracking procedure, and computational complexity are detailed in
Appendix~\ref{app:search-selection}.
The multiplier $r$ controls the complexity penalty under the counting
convention $p_k=2(k+1)$, and $10^{-12}$ provides a numerical floor.
The fit term uses $R_k$ without $\beta$ or the evidence reward.
Windowed evidence influences it through the locations in $B_k$. Accordingly, \(J\) and \(\mathrm{IC}_r\) serve distinct roles: \(J\) optimizes cut locations at a fixed \(k\), whereas \(\mathrm{IC}_r\) selects among candidates with different numbers of cuts. The global optimality of the dynamic program applies to \(J\) for each fixed k. 
The function $\operatorname{Delineate}$ performs the authorship
label and span delineation described below, returning sentence
labels, authorship boundaries, and predicted AI spans.



The selected count is
$\widehat{k}=\operatorname*{arg\,min}_{k\in\mathcal K}\mathrm{IC}_r(k)$,
with $\widehat B=B_{\widehat{k}}$.
Here, $\mathcal K$ contains the feasible counts searched under the
length constraint and the search cap $K_{\max}$, including zero.
Ties favor fewer cuts. Documents too short to admit multiple segments
are retained as a single statistical segment, including a fallback
when $n<L$.

\paragraph{Authorship Label  Assignment and Span Delineation.}
Given the selected statistical segmentation, we first assign authorship labels  to the resulting segments and then delineate contiguous human or
LLM spans. We assign each segment's arithmetic mean score to its
sentences, forming a piecewise-constant sequence $\boldsymbol z$.
We sort its distinct values and find the contiguous two-group split
that minimizes the within-group sum of squares, weighting each distinct
value equally. The threshold $\tau$ is the midpoint between the adjacent
values at this split. A sentence is labeled LLM if $z_i>\tau$ and human
otherwise. If $\boldsymbol z$ has only one distinct value, $\tau$ equals
that value and all sentences receive the human label.

Final authorship boundaries follow changes in the delineated labels: $\widehat C=
\{i\in\{1,\ldots,n-1\}:
\widehat y_i\neq\widehat y_{i+1}\}$. Adjacent statistical segments can receive the same label, so
$\widehat C\subseteq\widehat B$. Merging adjacent segments with identical
labels yields the maximal contiguous LLM spans returned as predicted
AI fragments. The deterministic thresholding, tie-breaking, and span-delineation rules are
provided in Appendix~\ref{app:source-delineation}. The complete inference procedure is summarized in Algorithm~\ref{alg:full-la-cpd-inference} in Appendix~\ref{app:full-inference}.

\section{Experiments}
\label{sec:experiments}


\paragraph{Datasets.}
We construct a GPT-5.5\footnote{Official model information is
available at \url{https://platform.openai.com/docs/models}.} human--LLM co-authored corpus from
WritingPrompts (WP)~\citep{fan2018hierarchical} and XSum
~\citep{narayan2018dont}, covering story and news texts, respectively.
The corpus includes human-only documents, documents with a single
authorship transition, and documents with multiple transitions. To
evaluate cross-model transfer and generalization, we further use
Claude Sonnet 4.5\footnote{Official model information is available at
\url{https://www.anthropic.com/news/claude-sonnet-4-5}.} datasets based on SQuAD~\citep{rajpurkar2016squad},
WP, and XSum, together with two public hybrid-text benchmarks:
the hybrid essays of~\citep{zeng2024boundary} and RoFT~\citep{dugan2020roft}. Details of dataset construction, corpus size,
data splits, and preprocessing are provided in
Appendix~\ref{app:dataset-construction}.

\paragraph{Baselines.}
We compare change-point detection methods for human--LLM authorship
boundaries (denoted by VCP and WCP~\citep{li2026segmenting}),
a sentence-level prediction baseline (denoted by SenPred~\citep{kushnareva2024roft}),
a majority-voting algorithm (denoted by Voting~\citep{zhang2024machine}),
TextTiling~\citep{hearst1997texttiling}, and a detector for
partially LLM-generated text (PaLD)~\citep{lei2025pald}.
WCP+AIC uses an AIC-style criterion to select the number
of statistical cuts. We also report LA-CPD oracle-$K$, which fixes the number of
statistical cuts to the reference number of authorship boundaries
while predicting cut locations and source labels, as a diagnostic
comparison. Implementation details and parameter settings for
all baselines are provided in Appendix~\ref{app:baselines}.

\paragraph{Parameter Settings and Evaluation Metrics.}
The local sampling-discrepancy detector uses Gemma-3-1B as its backbone
and is adapted with LoRA under sentence-level source supervision.
During inference, we set the segmentation parameters at
$(h,\alpha,\beta,r,K_{\max})=(6,0.75,1.25,2.5,3)$. We report sentence-level accuracy (Acc~\citep{kushnareva2024roft}), WindowDiff(WD;~\cite{pevzner-hearst-2002-critique}) and AI-fragment F1. All experiments were conducted on an NVIDIA GeForce RTX 4090 configuration with 24 GB of VRAM, using PyTorch~2.5.1 and CUDA~11.8. Training configurations, complete hyperparameter settings,
and metric definitions are provided in Appendix~\ref{app:parameter-settings}.
 





\subsection{Performance Comparison}
\label{sec:performance-comparison}

We evaluate the overall performance of LA-CPD on the GPT-5.5
human--LLM co-authored test set. All score-based methods, including SenPred, Voting, PaLD-scores, VCP, WCP, WCP+AIC, and LA-CPD, reuse the
same sentence-level score trajectory produced by a frozen local detector. TextTiling uses lexical cohesion to propose cut locations; its segment labels and evaluation scores are derived from the shared score trajectory. The results are reported in
Table~\ref{tab:main-results}.
\begin{table*}[t]
\centering
\small
\caption{Performance comparison on the GPT-5.5 human--LLM co-authored test set.
H0 reports accuracy, whereas mixed conditions report accuracy and AI-F1.
Best and second-best results among methods with unknown boundary counts
are shown in \textbf{bold} and \underline{underlined}, respectively.
Oracle-$K$, which uses the ground-truth boundary count, is reported only
for diagnostic purposes and is excluded from ranking.}
\label{tab:main-results}

\resizebox{\textwidth}{!}{%
\begin{tabular}{lcccccccc}
\toprule
\multirow{2}{*}{\textbf{Methods}}
& \multicolumn{3}{c}{Overall}
& \multicolumn{5}{c}{By condition} \\
\cmidrule(lr){2-4}
\cmidrule(lr){5-9}
& Acc $\uparrow$
& WD $\downarrow$
& AI-F1 $\uparrow$
& H0 Acc $\uparrow$
& B1 Acc / AI-F1 $\uparrow$
& B2 Acc / AI-F1 $\uparrow$
& B3-2 Acc / AI-F1 $\uparrow$
& B3-3 Acc / AI-F1 $\uparrow$ \\
\midrule

SenPred
& 0.598
& 0.874
& 0.086
& 0.404
& 0.690 / 0.086
& 0.696 / 0.091
& 0.527 / 0.072
& 0.673 / 0.135 \\

Voting
& 0.711
& 0.581
& 0.317
& 0.466
& \underline{0.836} / 0.419
& \textbf{0.829} / 0.402
& 0.642 / 0.208
& \textbf{0.784} / 0.434 \\

TextTiling
& 0.724
& 0.413
& 0.506
& 0.534
& \textbf{0.839} / 0.726
& 0.805 / 0.667
& 0.681 / \underline{0.334}
& \underline{0.760} / \textbf{0.567} \\

PaLD-scores
& 0.713
& 0.880
& 0.037
& 0.753
& 0.707 / 0.041
& 0.700 / 0.036
& \textbf{0.730} / 0.034
& 0.672 / 0.048 \\

VCP (MAD)
& 0.729
& 0.257
& 0.427
& \underline{0.891}
& 0.699 / 0.642
& 0.725 / 0.673
& 0.711 / 0.106
& 0.617 / 0.335 \\

WCP (MAD)
& 0.744
& \underline{0.248}
& 0.480
& 0.872
& 0.724 / 0.645
& 0.761 / 0.731
& 0.697 / 0.158
& 0.667 / 0.430 \\

WCP+AIC
& \underline{0.747}
& 0.270
& \underline{0.552}
& 0.786
& 0.793 / \underline{0.749}
& 0.787 / \underline{0.746}
& 0.633 / 0.280
& 0.737 / 0.550 \\

\midrule

\textbf{LA-CPD+AIC}
& \textbf{0.796}
& \textbf{0.231}
& \textbf{0.600}
& \textbf{0.907}
& 0.791 / \textbf{0.768}
& \underline{0.817} / \textbf{0.813}
& \underline{0.714} / \textbf{0.338}
& 0.750 / \underline{0.562} \\

LA-CPD oracle-$K$
& 0.866
& 0.186
& 0.736
& 1.000
& 0.910 / 0.945
& 0.908 / 0.935
& 0.714 / 0.456
& 0.797 / 0.679 \\

\bottomrule
\end{tabular}%
}
\end{table*}

\textbf{(1)} Experimental results show that, when the number of
boundaries is unknown, LA-CPD achieves the best overall performance on the GPT-5.5 test set. This observation indicates that, given the same local
detection scores, LA-CPD can exploit the overall sequential
structure among sentences and use local source predictions
to improve the delineation of contiguous AI spans.

\textbf{(2)} Current change-point detection methods enhanced
with AIC-style criteria, such as WCP+AIC, do not always
outperform conventional WCP. This is because changing the
cut-count selection rule affects the trade-off between these
two evaluation objectives. In comparison, the proposed LA-CPD
performs well overall across all metrics. This observation
indicates that the overall segmentation strategy of LA-CPD
can balance fragment detection and local boundary structure.

\textbf{(3)} WCP is an important baseline because it uses
length weighting to locate cut points for boundary detection
in human--LLM co-authored text with multiple boundaries.
LA-CPD achieves improved performance relative to this method.
This observation indicates that, building on weighted segment
fitting and cut-count selection, the overall LA-CPD framework,
which incorporates local transition information, can further
improve authorship segmentation.

\textbf{(4)} In addition, we fix the number of statistical
cuts to the ground-truth number of source boundaries, while
cut locations and source labels are still predicted by the
model. Overall detection performance improves under this
setting. This change indicates that ground-truth boundary-count
information helps delineate contiguous AI spans and that the
current cut-count selection procedure still leaves room
for improvement.

\subsection{Analysis of the Structured Segmentation Framework}
\label{sec:structured-segmentation-analysis}

The structured segmentation framework aims to organize sentence-level source
scores into contiguous authorship segments and to perform human--LLM
co-authored text detection with an unknown number of boundaries by jointly
optimizing cut-point locations and selecting the number of boundaries. To
further investigate its effects, we address the following questions: \textbf{RQ1:}  What contributions do the key components of the
structured segmentation framework make to authorship-span delineation
and boundary detection? \textbf{RQ2:} How well does the structured segmentation framework
generalize across different data distributions? \textbf{RQ3:} How stable is the structured segmentation framework
with respect to key segmentation and AIC parameters?
\begin{figure}[t]
    \centering
    \includegraphics[width=1\linewidth]{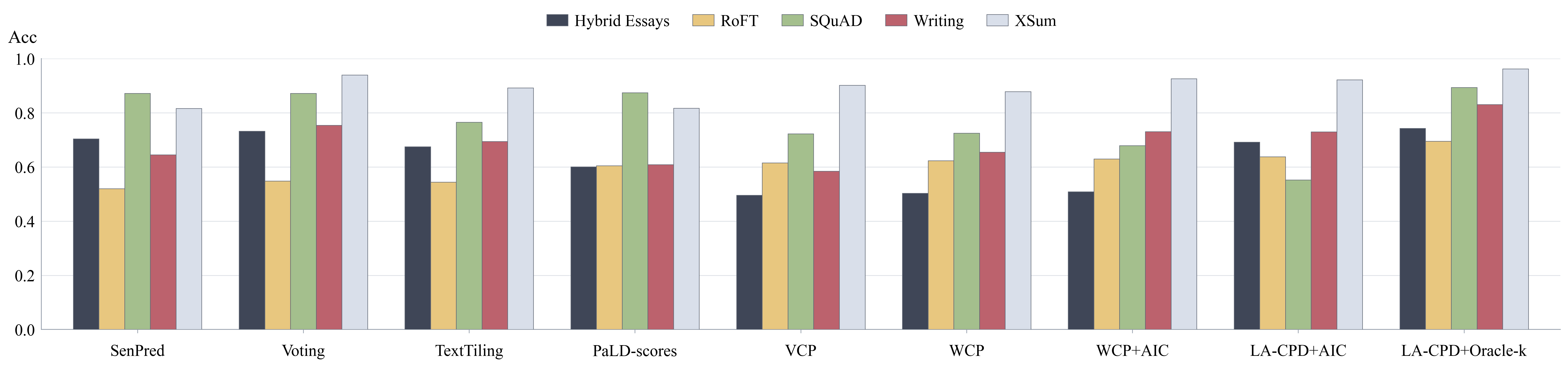}
    \caption{Cross-model generalization performance of LA-CPD in terms of
    sentence-level accuracy across different generation models.}
    \label{fig:generation-generalization}
\end{figure}
\paragraph{RQ1:}We conduct an ablation study on the structured-segmentation components to evaluate the  
\begin{wraptable}{r}{0.53\linewidth}
\vspace{-10pt}
\centering
\scriptsize
\setlength{\abovecaptionskip}{0pt}
\setlength{\belowcaptionskip}{2pt}
\setlength{\tabcolsep}{3pt}
\renewcommand{\arraystretch}{1.1}

\caption{Ablation study of LA-CPD.
Best and second-best results are shown in
\textbf{bold} and \underline{underlined}, respectively.}
\label{tab:ablation-results}

\resizebox{\linewidth}{!}{%
\begin{tabular}{lcccc}
\toprule
Methods
& Acc $\uparrow$
& WD $\downarrow$
& AI-F1 $\uparrow$
& B3-2 F1 $\uparrow$ \\
\midrule

LA-CPD (Full)
& \textbf{0.796}
& \underline{0.231}
& \underline{0.600}
& \textbf{0.338} \\

LA-CPD w/o $E(c;h)$
& 0.715
& 0.279
& 0.519
& 0.241 \\

LA-CPD w/o Length Weighting
& 0.781
& \textbf{0.223}
& 0.516
& 0.185 \\

LA-CPD w/o AIC
& \underline{0.789}
& 0.250
& \textbf{0.605}
& \underline{0.325} \\

\bottomrule
\end{tabular}%
}
\vspace{-10pt}
\end{wraptable}
impact of the structured segmentation framework on performance. The structured segmentation components of LA-CPD typically include local persistent transition evidence, length-weighted segment fitting, and AIC-style boundary-count selection. To verify its superiority, we compare LA-CPD+AIC with several variants, including those that
remove local persistent transition evidence, replace length weights with uniform weights, or remove AIC-style boundary-count selection. The experimental results are shown in Table~\ref{tab:ablation-results}.
Specifically, removing local persistent transition evidence leads to a noticeable deterioration across all metrics, indicating that aggregating persistent transition evidence over local score sequences helps improve authorship-span delineation and segmentation boundary structure. This observation and the ablation results indicate that the complete structured segmentation framework jointly contributes to authorship-span delineation and boundary detection. The complete ablation results are provided in
Appendix~\ref{app:full-ablation}.

Removing  windowed transition evidence causes a clear
degradation across all metrics, demonstrating the importance
of aggregating persistent local changes around candidate cuts.
The replacement of squared-length weights with uniform weights slightly
improves WD but substantially reduces AI-F1 and B3-2 F1,
indicating that length weighting is particularly beneficial for
AI-fragment delineation. Replacing AIC-based selection with the
MAD-based penalty slightly improves AI-F1 but degrades Acc and
WD, revealing a trade-off between fragment delineation and
boundary structure.


\paragraph{RQ2:}Different generation models, text domains, and datasets introduce
distribution shifts that can alter detector-score trajectories and
authorship-boundary patterns. To evaluate the applicability of the
structured segmentation framework under such shifts, we conduct
cross-model and cross-dataset/domain generalization experiments.
As shown in Figure~\ref{fig:generation-generalization}, LA-CPD+AIC shows its clearest advantage on RoFT, while its performance
is less consistent when generation models, text domains, and document
characteristics change simultaneously. These results suggest that the
quality of local detection scores and boundary-count selection remain
important limitations, leaving substantial room for improving the
cross-distribution generalization of LA-CPD. Complete generalization results are provided in Appendix ~\ref{app:generalization}.  

\paragraph{RQ3:}To evaluate the effects of key segmentation and AIC parameters on model performance, we conduct a hyperparameter sensitivity study on the GPT-5.5 human--LLM co-authored dataset. We vary the persistence window $h$, the transition-evidence weight $\alpha$, the segment-fitting weight $\beta$, and the complexity parameter $r$ individually, while keeping all other settings fixed. The results are reported in Table~\ref{tab:hyperparameter-sensitivity}. The results show that the parameters affect performance differently. As the persistence window increases, WD first decreases and then
increases, indicating that moderate expansion of the local evidence aggregation range can provide more stable authorship-transition information. In contrast, varying $\alpha$ and $\beta$ within the tested ranges causes only small performance fluctuations, suggesting that the structured segmentation framework is reasonably robust to these two weighting parameters. The AIC penalty coefficient $r$ introduces a trade-off among the evaluation objectives, and an excessively strong complexity penalty may impair AI-fragment delineation and overall segmentation quality.

\begin{table*}[t]
\centering
\scriptsize
\setlength{\tabcolsep}{4pt}
\renewcommand{\arraystretch}{1.15}
\caption{Hyperparameter sensitivity of LA-CPD on the GPT-5.5
human--LLM co-authored dataset. $\dagger$ denotes the default
configuration. Best and second-best results within each parameter
group are shown in \textbf{bold} and \underline{underlined},
respectively.}
\label{tab:hyperparameter-sensitivity}
\resizebox{\textwidth}{!}{%
\begin{tabular}{l*{14}{c}}
\toprule
\multicolumn{1}{c}{\multirow{2}{*}{\textbf{Metric}}}
& \multicolumn{4}{c}{$h$}
& \multicolumn{3}{c}{$\alpha$}
& \multicolumn{3}{c}{$\beta$}
& \multicolumn{4}{c}{$r$} \\
\cmidrule(lr){2-5}
\cmidrule(lr){6-8}
\cmidrule(lr){9-11}
\cmidrule(lr){12-15}
& 1 & 3 & $6^{\dagger}$ & 8
& 0.25 & $0.75^{\dagger}$ & 1.5
& 0.75 & $1.25^{\dagger}$ & 2.0
& 1.0 & 2.0 & $2.5^{\dagger}$ & 3.0 \\
\midrule
Acc $\uparrow$
& 0.702 & 0.766 & \textbf{0.796} & \underline{0.793}
& \underline{0.794} & \textbf{0.796} & 0.786
& 0.790 & \textbf{0.796} & \underline{0.793}
& 0.769 & 0.795 & \textbf{0.796} & 0.786 \\

AI-F1 $\uparrow$
& 0.494 & 0.561 & \textbf{0.600} & \underline{0.567}
& \underline{0.592} & \textbf{0.600} & 0.575
& 0.577 & \textbf{0.600} & \underline{0.588}
& \textbf{0.614} & \underline{0.607} & 0.600 & 0.544 \\

WD $\downarrow$
& 0.301 & 0.277 & \textbf{0.231} & \underline{0.247}
& \underline{0.255} & \textbf{0.231} & 0.259
& 0.257 & \textbf{0.231} & \underline{0.255}
& 0.321 & 0.267 & \textbf{0.231} & \underline{0.244} \\
\bottomrule
\end{tabular}%
}
\end{table*}

\section{CONCLUSION}
\label{sec:conclusion & limitations}

In this work, we propose Local-Evidence-Aware Change-Point Detection
(LA-CPD) to delineate contiguous human- and LLM-authored spans in
human--LLM co-authored documents with unknown boundary locations and
counts. Experimental results show that persistent local transition
evidence is a primary factor in improving boundary localization,
whereas length weighting promotes the delineation of contiguous
AI-authored spans. By introducing structured change-point detection
into human--LLM co-authored text analysis, this work provides a new
perspective for boundary detection in human--AI collaboration.

\textbf{Limitations}
LA-CPD depends on the discriminative ability of the local detector,
and cross-distribution detection remains challenging. Dynamic
programming also introduces computational overhead for long documents.
Future work will explore more robust detectors, finer-grained
authorship localization, and more efficient inference.

\subsection*{AI use statement}

In this work, we used GPT-5.5 and Claude Sonnet 4.5 to construct human--LLM co-authored corpora and evaluate cross-model generalization. The generative models are used only to produce the LLM-authored segments in the experimental data; document-level source labels are determined by the data-construction procedure rather than by the generative models themselves. Detailed prompt templates, model configurations, data filtering, and data-splitting procedures are provided in the appendix. The authors are responsible for data processing, experimental analysis, interpretation of the results, and the accuracy of the final manuscript.

\subsection*{Ethics statement}

This study does not involve human-subject experiments, user surveys, intervention studies, or the collection of new human-subject data. We use publicly available datasets, including WritingPrompts, XSum, SQuAD, RoFT, and Hybrid Essays, in accordance with their respective licenses and terms of use. We neither identify nor infer the personal identities of text authors, nor use private or sensitive texts for training or evaluation.
This work aims to improve provenance analysis, content attribution, and accountability assessment in human--LLM co-authored text. Because detection performance may be affected by the generation model, text domain, document length, and textual content, LA-CPD may produce false positives and false negatives. Therefore, its outputs should not be used as the sole basis for academic sanctions, copyright decisions, hiring decisions, or legal liability determinations, but should be considered alongside human review and other evidence. We do not advocate using the proposed method for unauthorized individual surveillance or fully automated punitive decisions.

\bibliography{iclr2027_conference}
\bibliographystyle{iclr2027_conference}

\clearpage
\appendix
\startcontents[appendix]

\begingroup
\normalfont\normalsize
\setlength{\parindent}{0pt}
\setlength{\parskip}{0.5pc}
\hypersetup{hidelinks}

\section*{Appendix}
\subsection*{Table of Contents}

\titlecontents{section}[2em]
  {\addvspace{0.6em}\normalfont\normalsize\mdseries\scshape}
  {\contentslabel{1.5em}}
  {}
  {\hfill\normalfont\normalsize\contentspage}

\titlecontents{subsection}[4em]
  {\normalfont\normalsize\mdseries}
  {\contentslabel{2.5em}}
  {}
  {\titlerule*[0.7em]{.}\contentspage}

\printcontents[appendix]{}{1}{\setcounter{tocdepth}{2}}

\endgroup
\clearpage

\section{Method Details and Implementation}
\label{app:method-details}

\subsection{Local Detector Adaptation}
\label{app:detector-training}

The local detector consists of an adaptable scoring model and a reference model with fixed parameters. Both models use
\texttt{google/gemma-3-1b-pt} as the backbone. The scoring model is adapted with LoRA-based authorship supervision. The main experiment continues adaptation from an existing LoRA checkpoint. The configuration is summarized in Table~\ref{tab:detector-training-config}.

\begin{table}[h]
\centering
\caption{Configuration of the continued local-detector adaptation.}
\label{tab:detector-training-config}
\small
\begin{tabular}{p{0.43\linewidth}p{0.48\linewidth}}
\toprule
Component & Setting \\
\midrule
Scoring and reference backbone & Gemma-3-1B \\
LoRA target modules &
\texttt{q\_proj}, \texttt{k\_proj}, \texttt{v\_proj}, \texttt{o\_proj} \\
LoRA rank & 4 \\
LoRA scaling parameter & 16 \\
LoRA dropout & 0.05 \\
Continued-adaptation training pairs & 4,000 \\
Optimizer & AdamW \\
Learning rate & $10^{-4}$ \\
Training pairs per batch & 1 \\
Configured number of epochs & 1 \\
Random seed & 42 \\
\bottomrule
\end{tabular}
\end{table}

Let $\{(u_{\mathrm H}^{(b)},u_{\mathrm L}^{(b)})
\}_{b=1}^{M}$ denote a training batch, where the two inputs originate from human and LLM
sources, respectively. The adaptation objective is the mean-score separation
loss
\begin{equation}
\mathcal{L}_{\mathrm{sep}}
=
\frac{1}{M}
\sum_{b=1}^{M}
\left[
d_{\theta}\!\left(u_{\mathrm H}^{(b)}\right)
-
d_{\theta}\!\left(u_{\mathrm L}^{(b)}\right)
\right].
\label{eq:app-separation-loss}
\end{equation}

The data loader forms training pairs using the same index in the two source
lists. Minimizing Eq.~(\ref{eq:app-separation-loss}) encourages higher mean
scores for LLM text than for human text. The loss does not use document cut
locations. Only the scoring model's LoRA parameters are updated during
adaptation; the scoring backbone and the reference model remain frozen.
Training uses AdamW and automatic mixed precision.

After adaptation, the scoring model is frozen. Writing its adapted parameters
as $\widehat{\theta}$, we define $\phi(x)=d_{\widehat{\theta}}(x).$ For a document $\mathcal{D}=(u_1,\ldots,u_n)$, the sentence-level score
trajectory is $\boldsymbol{q} = \left(
\phi(u_1),\ldots,\phi(u_n)
\right).$

\subsection{Complete Definition of the Local Score}
\label{app:score-computation}

The local score is computed from the standardized deviation of the
observed-token log probability from its conditional expectation under a
frozen reference distribution. Let
$x=(x_1,\ldots,x_T)$ denote the encoded token sequence of a sentence,
and let $t=2,\ldots,T$ index the prediction positions. For a vocabulary
token $v\in\mathcal{V}$, define
\begin{equation}
a_{\theta,t}(v)
=
\log p_{\theta}(v\mid x_{<t}),
\qquad
\pi_t(v)
=
p_{\mathrm{ref}}(v\mid x_{<t}),
\label{eq:app-score-prob}
\end{equation}
where $p_{\theta}$ is the scoring model and
$p_{\mathrm{ref}}$ is the frozen reference model.

The conditional mean and variance of the scoring-model log probability
under the reference distribution are computed by summing over the
vocabulary:
\begin{equation}
\mu_{\theta,t}(x)
=
\sum_{v\in\mathcal{V}}
\pi_t(v)a_{\theta,t}(v),
\label{eq:app-score-mean}
\end{equation}
\begin{equation}
\sigma_{\theta,t}^{2}(x)
=
\sum_{v\in\mathcal{V}}
\pi_t(v)
\left(
a_{\theta,t}(v)-\mu_{\theta,t}(x)
\right)^2.
\label{eq:app-score-var}
\end{equation}

For consistency with the notation in
Eq.~(\ref{eq:separation-loss}), we define the score-model and
reference-model representations of a sentence as
\begin{equation}
\mathcal{M}_{\mathrm{score}}(x)
=
\left\{
a_{\theta,t}(x_t)
\right\}_{t=2}^{T},
\qquad
\mathcal{M}_{\mathrm{ref}}(x)
=
\left\{
\left(
\mu_{\theta,t}(x),
\sigma_{\theta,t}^{2}(x)
\right)
\right\}_{t=2}^{T}.
\label{eq:app-model-representations}
\end{equation}

The sampling-discrepancy function
$d_{\theta}\!\left(
\mathcal{M}_{\mathrm{score}}(x),
\mathcal{M}_{\mathrm{ref}}(x)
\right)$ is abbreviated as $d_{\theta}(x)$ below and is defined by
\begin{equation}
d_{\theta}(x)
=
\frac{
\displaystyle
\sum_{t=2}^{T}
\left[
a_{\theta,t}(x_t)-\mu_{\theta,t}(x)
\right]
}{
\displaystyle
\sqrt{
\max\left\{
\sum_{t=2}^{T}
\sigma_{\theta,t}^{2}(x),
\varepsilon_{\mathrm{var}}
\right\}
}
},
\label{eq:app-discrepancy-score}
\end{equation}
where $\varepsilon_{\mathrm{var}}=10^{-4}$ is a numerical floor applied
to the accumulated conditional variance.

This computation follows the sampling-discrepancy formulation of
Fast-DetectGPT~\citep{bao2024fast}. The conditional moments are obtained
through analytical vocabulary summation, so no explicit sampling of
replacement tokens is required. During inference, the resulting
statistic is used directly as the sentence-level authorship score.

\subsection{Interval Statistics and Windowed Transition Evidence}
\label{app:interval-statistics}

Interval fitting uses normalized squared token-length weights. The
sentence length $\ell_i$ is computed using the scoring tokenizer without
adding special tokens. We define
\begin{equation}
\label{eq:app-length-weight}
w_i =
\frac{n\ell_i^2}{\sum_{j=1}^{n}\ell_j^2},
\qquad
\sum_{i=1}^{n} w_i = n.
\end{equation} This normalization preserves the relative contribution induced by
sentence length while making the average document weight equal to one.
The length used in Eq.~(\ref{eq:app-length-weight}) is computed
separately from the encoded sequence length used in
Appendix~\ref{app:score-computation}.

We compute interval residuals using three prefix statistics:
\begin{equation}
P_w(t)=\sum_{i=1}^{t}w_i,
\qquad
P_q(t)=\sum_{i=1}^{t}w_iq_i,
\qquad
P_{q^2}(t)=\sum_{i=1}^{t}w_iq_i^2.
\label{eq:app-prefix-statistics}
\end{equation}
All three statistics are initialized to zero at $t=0$. For an interval
$(s,e]$, define
\begin{equation}
W_{s:e}=P_w(e)-P_w(s),
\qquad
Q_{s:e}=P_q(e)-P_q(s),
\qquad
S_{s:e}=P_{q^2}(e)-P_{q^2}(s).
\label{eq:app-interval-statistics}
\end{equation}

The weighted mean and interval residual are
\begin{equation}
\bar q_{s:e}
=
\frac{Q_{s:e}}{W_{s:e}},
\qquad
V(s,e)
=
S_{s:e}
-
\frac{Q_{s:e}^{\,2}}{W_{s:e}}.
\label{eq:app-segment-residual}
\end{equation}

The prefix statistics are constructed in $O(n)$ time, after which each
interval residual can be obtained in $O(1)$ time. The implementation clips
negative residuals caused by numerical roundoff to zero. Empty intervals and
intervals whose total weight is no greater than $10^{-12}$ are treated as
infeasible. Valid interval costs are cached before dynamic programming.

The windowed evidence follows a local residual decomposition identity. For a
candidate cut $c$, define $a=\max(0,c-h), b=\min(n,c+h).$

The left and right windows are $(a,c]$ and $(c,b]$, with weighted means
$\mu_L$ and $\mu_R$ and total weights $W_L$ and $W_R$, respectively. Their
combined weighted mean is
\begin{equation}
\mu
=
\frac{W_L\mu_L+W_R\mu_R}
{W_L+W_R}.
\label{eq:app-window-mean}
\end{equation}

For the left window,
\begin{equation}
\begin{aligned}
\sum_{i=a+1}^{c}w_i(q_i-\mu)^2
={}&
\sum_{i=a+1}^{c}w_i(q_i-\mu_L)^2
+
W_L(\mu_L-\mu)^2.
\end{aligned}
\label{eq:app-left-window-decomposition}
\end{equation}

The cross term vanishes because $\sum_{i=a+1}^{c}
w_i(q_i-\mu_L)=0.$ The same identity holds for the right window. Therefore,
\begin{equation}
\begin{aligned}
V(a,b)-V(a,c)-V(c,b)
&=
W_L(\mu_L-\mu)^2
+
W_R(\mu_R-\mu)^2\\
&=
\frac{W_LW_R}{W_L+W_R}
(\mu_L-\mu_R)^2\\
&=
E(c;h).
\end{aligned}
\label{eq:app-window-evidence-identity}
\end{equation}

Thus, the windowed evidence equals the reduction in weighted residual obtained
by replacing one common score level with two independent local levels. When
$h=1$, the two windows contain only the sentences immediately adjacent to the
candidate cut, and
\begin{equation}
E(c;1)
=
\frac{w_cw_{c+1}}
{w_c+w_{c+1}}
(q_c-q_{c+1})^2.
\label{eq:app-window-h1}
\end{equation}
For $h>1$, the evidence aggregates multiple local observations through the
weighted means on both sides. The implementation computes $E(c;h)$ for every
inter-sentence position, after which the complete partition objective and
segment-length constraints determine the selected statistical cuts.

\subsection{Search Space, Dynamic Programming, and Model Selection}
\label{app:search-selection}

Assume a non-empty document. Let $h\geq1$ and $K_{\max}\geq0$ be integers. $L=\max\{1,L_{\min}\}.$ When $\alpha>0$, we further set $L=\max\{L,h\}.$ If $n<L$, the method retains an empty statistical cut set and performs single-segment authorship delineation on the document. For $n\geq L$, the searched cut-count set is
\[
\mathcal{K}=\{0,\ldots,K_{\mathrm{eff}}\},
\]
where
\begin{equation}
K_{\mathrm{eff}}
=
\min\left\{
K_{\max},
\max\left\{
0,
\left\lfloor
\frac{n}{\max\{2,L\}}
\right\rfloor-1
\right\}
\right\}.
\label{eq:app-keff}
\end{equation}
This expression limits the number of cut counts searched by combining
the user-specified cap $K_{\max}$ with a length-dependent search cap.
The minimum length of each statistical segment is still enforced
separately by the dynamic-programming constraint that every segment
contains at least $L$ sentences.
This range accounts for both the minimum segment length and the additional
search bound used by the implementation. The shared dynamic program stores the best objective value and its predecessor. Let $F_k(e)$ denote the minimum candidate objective for the first $e$ sentences with exactly $k$ statistical cuts. The initialization is
\begin{equation}
F_0(e)
=
\beta V(0,e),
\qquad
e\geq L,
\label{eq:app-dp-init}
\end{equation}
with infeasible states assigned $+\infty$. For $k\geq1$ and
$e\geq(k+1)L$, the recurrence is
\begin{equation}
F_k(e)
=
\min_{\substack{
t\in\mathbb{Z}\\
kL\leq t\leq e-L
}}
\left\{
F_{k-1}(t)
+
\beta V(t,e)
-
\alpha E(t;h)
\right\}.
\label{eq:app-dp}
\end{equation}

The windowed evidence is computed once for the full document and remains fixed
across dynamic-programming states. Candidate predecessors are enumerated in
increasing order. When two candidates have the same objective value, the
smaller predecessor is retained.

The recurrence gives the global optimum of the candidate objective $J$ for
each fixed $k$. Any feasible partition has a final cut $t$, so its objective
decomposes into a prefix objective, the residual of the final segment, and the
reward associated with the final cut. Once $t$ is fixed, the latter two terms
are fixed. If the prefix is not optimal for state $F_{k-1}(t)$, replacing it
with the optimal prefix reduces the complete objective while preserving the
cut count and all length constraints. Induction from the zero-cut state and
enumeration of all feasible final cuts therefore yield the global minimum for
each fixed $k$.

For each $k\in\mathcal{K}$, the stored predecessors are backtracked from
$(k,n)$ to obtain an ordered candidate cut set $B_k=\{b_1^{(k)},\ldots,b_k^{(k)}\}.$
After adding the endpoints $b_0^{(k)}=0, b_{k+1}^{(k)}=n,$ the original weighted residual is recomputed as
\begin{equation}
R_k
=
\sum_{j=0}^{k}
V\!\left(
b_j^{(k)},
b_{j+1}^{(k)}
\right).
\label{eq:app-candidate-residual}
\end{equation}

The cut count is selected using the complexity-counting convention $p_k=2(k+1),$ and the AIC-style criterion
\begin{equation}
\mathrm{IC}_r(k)
=
n\log\left(
\max\left\{
\frac{R_k}{n},
10^{-12}
\right\}
\right)
+
2rp_k.
\label{eq:app-aic}
\end{equation}
Equivalently, the complexity term is $4r(k+1)$. The selected count is
\begin{equation}
\widehat{k}
=
\min\left(
\operatorname*{arg\,min}_{k\in\mathcal{K}}
\mathrm{IC}_r(k)
\right),
\qquad
\widehat{B}=B_{\widehat{k}},
\label{eq:app-model-selection}
\end{equation}
where the outer minimum implements the tie-breaking rule that favors fewer
cuts.

The original residual and the dynamic-programming objective serve different
purposes. For a dedlineated candidate,
\begin{equation}
F_k(n)
=
\beta R_k
-
\alpha
\sum_{c\in B_k}
E(c;h).
\label{eq:app-dp-residual-relation}
\end{equation}
Thus, the AIC-style criterion uses $R_k$ without the factor $\beta$ and without
the transition-evidence reward. Windowed evidence affects the criterion
indirectly by changing the candidate cut locations and their corresponding
residuals. The fixed-$k$ global optimality claim applies to $J$, whereas
$\mathrm{IC}_r$ compares the candidates obtained for different cut counts.

The main experiment uses $(h,\alpha,\beta,r,K_{\max},L_{\min}) = (6,0.75,1.25,2.5,3,1),$ which gives an effective minimum segment length of $L=6$.

The computational cost is summarized in Table~\ref{tab:app-complexity}.

\begin{table}[t]
\centering
\caption{Time and storage costs of structured segmentation, excluding detector
forward passes.}
\label{tab:app-complexity}
\small
\begin{tabular}{lcc}
\toprule
Operation & Time & Main storage \\
\midrule
Prefix statistics & $O(n)$ & $O(n)$ \\
Interval residual cache & $O(n^2)$ & $O(n^2)$ \\
Windowed evidence & $O(nh)$ & $O(n)$ \\
Shared dynamic programming &
$O((K_{\mathrm{eff}}+1)n^2)$ &
$O((K_{\mathrm{eff}}+1)n)$ \\
Candidate backtracking and residual accumulation &
$O((K_{\mathrm{eff}}+1)^2)$ &
Does not change the overall bound \\
Authorship delineation &
$O(n\log n+m^2)$ &
$O(n)$ \\
\bottomrule
\end{tabular}
\end{table}

Here, $m$ denotes the number of distinct values in the piecewise-constant
score sequence, satisfying $m\leq \widehat{k}+1\leq n.$ Excluding detector forward passes, the overall time and space complexities are
$
O\!\left(
(K_{\mathrm{eff}}+1)n^2+nh
\right)
$
and
$
O\!\left(
n^2+(K_{\mathrm{eff}}+1)n
\right),
$
respectively.

\subsection{Deterministic Authorship Label and Span Delineation}
\label{app:source-delineation}

Authorship delineation uses the arithmetic mean score of each selected statistical
segment. For a selected segment
$(\widehat b_j,\widehat b_{j+1}]$, define
\begin{equation}
z_i
=
\frac{
\displaystyle
\sum_{t=\widehat b_j+1}^{\widehat b_{j+1}}q_t
}{
\widehat b_{j+1}-\widehat b_j
},
\qquad
\widehat b_j<i\leq\widehat b_{j+1}.
\label{eq:app-piecewise-score}
\end{equation}
This step uses an unweighted arithmetic mean and is separate from the
length-weighted residual used for candidate segmentation.

Let the distinct values in $\boldsymbol{z}$ be $
\{v_1,\ldots,v_m\}
=
\operatorname{unique}(\boldsymbol{z}),
\qquad
v_1<\cdots<v_m.
$ When $m\geq2$, for $p=1,\ldots,m-1$, define
\begin{equation}
\bar v_{1:p}
=
\frac{1}{p}
\sum_{j=1}^{p}v_j,
\qquad
\bar v_{p+1:m}
=
\frac{1}{m-p}
\sum_{j=p+1}^{m}v_j,
\label{eq:app-cluster-means}
\end{equation}
and
\begin{equation}
H(p)
=
\sum_{j=1}^{p}
(v_j-\bar v_{1:p})^2
+
\sum_{j=p+1}^{m}
(v_j-\bar v_{p+1:m})^2.
\label{eq:app-cluster-objective}
\end{equation}

The selected split is
\begin{equation}
p^\star
=
\min\left(
\operatorname*{arg\,min}_{1\leq p<m}
H(p)
\right),
\qquad
\tau
=
\frac{
v_{p^\star}+v_{p^\star+1}
}{2}.
\label{eq:app-threshold}
\end{equation}
Each distinct score value contributes once to the clustering objective; its
frequency among sentences or statistical segments is not used as an
additional weight.

When $m=1$, the fallback threshold is $
\tau=v_1.
$ The authorship label is then assigned by
\begin{equation}
\widehat{y}_i
=
\begin{cases}
\mathrm{L}, & z_i>\tau,\\
\mathrm{H}, & z_i\leq\tau.
\end{cases}
\label{eq:app-source-label}
\end{equation}
Thus, a sequence with only one distinct score level is assigned the human
label throughout, including the zero-cut case.

The final authorship boundaries are obtained from changes in the delineated
labels: $\widehat{C}
=
\left\{
i\in\{1,\ldots,n-1\}:
\widehat{y}_i\neq\widehat{y}_{i+1}
\right\}.$
All sentences within one statistical segment have the same score level and
therefore the same label. Adjacent statistical segments may receive the same
label, which gives $\widehat{C}\subseteq\widehat{B}.$ After merging adjacent segments with identical labels, all maximal contiguous
segments labeled LLM are returned as predicted AI spans.

\subsection{Full LA-CPD Inference Algorithm}
\label{app:full-inference}

Algorithm~\ref{alg:full-la-cpd-inference} summarizes the complete
inference procedure described in Sections~A.3--A.5. It covers candidate
construction, fixed-count dynamic programming, AIC-style cut-count
selection, and deterministic authorship label and span delineation.

\begin{algorithm}[H]
\small
\DontPrintSemicolon
\SetAlgoLined
\SetAlgoNlRelativeSize{-1}
\SetAlCapSkip{0.5em}

\caption{Full LA-CPD Inference}
\label{alg:full-la-cpd-inference}

\KwIn{Document $\mathcal D=(u_1,\ldots,u_n)$, frozen detector $\phi$,
and parameters
$\Theta=(h,\alpha,\beta,r,K_{\max},L_{\min})$}

\KwOut{Sentence labels $\widehat{\boldsymbol y}$, authorship boundaries
$\widehat{\mathcal C}$, and predicted AI spans
$\widehat{\mathcal S}_{\mathrm{AI}}$}

Compute sentence scores $q_i\leftarrow\phi(u_i)$ and sentence-length
weights $\boldsymbol w$\;

Determine the effective minimum length $L$ and feasible cut-count set
$\mathcal K$\;

\If{$n<L$}{
    \KwRet{$\operatorname{Delineate}(\boldsymbol q,\varnothing)$}\;
}

\BlankLine
Compute segment residuals $V(s,e)$ and windowed transition evidence
$E(c;h)$ using Eqs.~(\ref{eq:segment-residual})--%
(\ref{eq:windowed-evidence})\;

Initialize $F_0(e)=\beta V(0,e)$ for $e\geq L$ and set infeasible
states to $+\infty$\;

\BlankLine
\ForEach{$k\in\mathcal K\setminus\{0\}$}{
    Use Eq.~(\ref{eq:dp}) to obtain the optimal fixed-count
    partition $B_k$ and store its predecessor pointers\;
}

\BlankLine
\ForEach{$k\in\mathcal K$}{
    \eIf{$k=0$}{
        $B_0\leftarrow\varnothing$\;
    }{
        Backtrack the predecessor pointers to obtain $B_k$\;
    }

    Recompute
    $R_k\leftarrow\sum_{j=0}^{k}
    V(b_j^{(k)},b_{j+1}^{(k)})$\;

    Evaluate $\mathrm{IC}_r(k)$ using Eq.~(\ref{eq:aic-criterion})\;
}

Select $\widehat{k}=\arg\min_{k\in\mathcal K}\mathrm{IC}_r(k)$,
favoring fewer cuts in a tie, and set
$\widehat B\leftarrow B_{\widehat{k}}$\;

\BlankLine
Construct the statistical segments induced by $\widehat B$\;

Assign the arithmetic mean score of each selected segment to all
sentences in that segment, obtaining the piecewise-constant sequence
$\boldsymbol z$\;

Sort the distinct values of $\boldsymbol z$ and determine the
two-group split that minimizes the within-group sum of squares\;

Set $\tau$ to the midpoint between the two values at the selected split,
and assign LLM labels when $z_i>\tau$ and human labels otherwise\;

\BlankLine
Set
\[
\widehat{\mathcal C}\leftarrow
\{i\in\{1,\ldots,n-1\}:
\widehat y_i\neq\widehat y_{i+1}\};
\]

Merge adjacent segments with identical labels and return
$(\widehat{\boldsymbol y},\widehat{\mathcal C},
\widehat{\mathcal S}_{\mathrm{AI}})$\;

\end{algorithm}

\section{Experimental Details}
\label{app:experimental-details}

\subsection{Dataset Settings}
\label{app:dataset-construction}

\paragraph{GPT-5.5 human--LLM co-authored corpus.}
We collect human texts from WritingPrompts (WP) and XSum and assign
a \texttt{source\_family\_id} to each source document. WP provides
stories, whereas XSum provides news articles. We segment the texts
using the spaCy English sentencizer and count tokens using the
\texttt{cl100k\_base} encoding in \texttt{tiktoken}. Authorship spans
are represented as half-open character intervals
$[\mathrm{start},\mathrm{end})$.

The corpus contains five conditions, where $\mathrm{H}$ and
$\mathrm{L}$ denote human and LLM authorship, respectively:
\begin{itemize}[leftmargin=*,itemsep=2pt,topsep=4pt]
    \item \textbf{H0:}
    The entire document is human-written and contains no AI span.

    \item \textbf{B1:}
    An LLM continues a human-written prefix from a sentence boundary.

    \item \textbf{B2:}
    The human-written prefix ends within a sentence, which the LLM
    continues.

    \item \textbf{B3-2:}
    The document contains two authorship transitions, following
    $\mathrm{H}\rightarrow\mathrm{L}\rightarrow\mathrm{H}$.

    \item \textbf{B3-3:}
    The document contains three authorship transitions, following
    $\mathrm{H}\rightarrow\mathrm{L}\rightarrow
    \mathrm{H}\rightarrow\mathrm{L}$.
\end{itemize}

An entirely AI-written condition, A0, was considered during corpus
design but is absent from the generated corpus and the reported
experiments. It is therefore excluded from the corpus statistics.

H0 examples retain the human-written text without invoking a
generation model. For B1, B2, and B3, GPT-5.5 generates AI spans from
the supplied human context. B1 starts the continuation with a new
sentence, whereas B2 continues an unfinished sentence. B3-2 fills
a missing span between left and right human contexts. B3-3 first
fills an intermediate AI span and then generates a final
continuation from the assembled prefix. All prompts instruct the
model to return only the newly generated text, avoid repeating
the context, and preserve its topic, style, tone, and language. The complete templates are provided below.

\begin{lacpdpromptbox}{
    Prompt-based Generation: Shared System Prompt
}
\lacpdpromptheading{System prompt}

You are a creative writing assistant. Continue the given
human-written text naturally and fluently, matching the topic,
style, tone, and language of what came before. Return only the
requested text. Do not explain, label, summarize, or repeat the
given context. Follow the requested word-count range carefully.
\end{lacpdpromptbox}

\begin{lacpdpromptbox}{
    Prompt-based Generation: Single-transition Templates
}
\lacpdpromptheading{B1: Continuation from a sentence boundary}

Continue the following text from the next sentence.

Requirements:
\begin{lacpdpromptitems}
    \item Write only the continuation.
    \item Do not repeat any part of the prefix.
    \item Start naturally as a new sentence.
    \item Continue the same topic, style, tone, and language
    as the prefix.
    \item Write \texttt{\{min\_words\}} to
    \texttt{\{max\_words\}} words. Aim for approximately
    \texttt{\{target\_words\}} words.
    \item The word-count requirement is important.
\end{lacpdpromptitems}

Prefix:\par
\texttt{\{human\_prefix\}}

Continuation:

\lacpdpromptdivider

\lacpdpromptheading{B2: Continuation from within a sentence}

Continue the following text from the exact point where it stops.

Requirements:
\begin{lacpdpromptitems}
    \item Write only the continuation.
    \item Do not repeat any part of the prefix.
    \item The prefix ends in the middle of a sentence.
    \item Continue the unfinished sentence naturally.
    \item Do not restart the sentence.
    \item Continue the same topic, style, tone, and language
    as the prefix.
    \item Write \texttt{\{min\_words\}} to
    \texttt{\{max\_words\}} words. Aim for approximately
    \texttt{\{target\_words\}} words.
    \item The word-count requirement is important.
\end{lacpdpromptitems}

Prefix:\par
\texttt{\{human\_prefix\}}

Continuation:
\end{lacpdpromptbox}

\clearpage
\begin{lacpdpromptbox}{
    Prompt-based Generation: B3-2 Templates
}
\lacpdpromptheading{
    Variant 1: Enter AI within a sentence;
    return to human at a sentence boundary
}

Fill in the missing middle segment of the following hybrid text.

Requirements:
\begin{lacpdpromptitems}
    \item Return only the missing segment.
    \item Do not repeat the left or right context.
    \item The left context ends in the middle of a sentence.
    \item Start by continuing the unfinished sentence naturally.
    \item The missing segment must end as a complete sentence.
    \item The right context begins as a new human-written sentence
    after the missing segment.
    \item Match the topic, style, tone, and language of both contexts.
    \item Write \texttt{\{min\_words\}} to
    \texttt{\{max\_words\}} words. Aim for approximately
    \texttt{\{target\_words\}} words.
    \item The word-count requirement is important.
\end{lacpdpromptitems}

Left context:\par
\texttt{\{human\_left\_prefix\}}

Right context:\par
\texttt{\{human\_right\_suffix\}}

Missing segment:

\lacpdpromptdivider

\lacpdpromptheading{
    Variant 2: Enter AI at a sentence boundary;
    return to human within a sentence
}

Fill in the missing middle segment of the following hybrid text.

Requirements:
\begin{lacpdpromptitems}
    \item Return only the missing segment.
    \item Do not repeat the left or right context.
    \item The left context ends with a complete sentence.
    \item Start naturally as a new sentence.
    \item The missing segment must end in the middle of a sentence
    so that the right context can continue it naturally.
    \item Do not restart the sentence that the right context continues.
    \item Match the topic, style, tone, and language of both contexts.
    \item Write \texttt{\{min\_words\}} to
    \texttt{\{max\_words\}} words. Aim for approximately
    \texttt{\{target\_words\}} words.
    \item The word-count requirement is important.
\end{lacpdpromptitems}

Left context:\par
\texttt{\{human\_left\_prefix\}}

Right context:\par
\texttt{\{human\_right\_suffix\}}

Missing segment:
\end{lacpdpromptbox}

\begin{lacpdpromptbox}{
    Prompt-based Generation: B3-3 Templates
}
\lacpdpromptheading{Step 1: Generate the first AI span}

Fill in the missing first AI-written segment of the following
hybrid text.

Requirements:
\begin{lacpdpromptitems}
    \item Return only the missing segment.
    \item Do not repeat the left or right context.
    \item The left context ends in the middle of a sentence.
    \item Start by continuing the unfinished sentence naturally.
    \item The missing segment must end as a complete sentence.
    \item The right context begins as a new human-written sentence
    after the missing segment.
    \item Match the topic, style, tone, and language of both contexts.
    \item Write \texttt{\{ai\_segment\_1\_min\_words\}} to
    \texttt{\{ai\_segment\_1\_max\_words\}} words.
    Aim for approximately
    \texttt{\{ai\_segment\_1\_target\_words\}} words.
    \item The word-count requirement is important.
\end{lacpdpromptitems}

Left context:\par
\texttt{\{human\_part\_1\}}

Right context:\par
\texttt{\{human\_part\_2\}}

Missing segment:

\lacpdpromptdivider

\lacpdpromptheading{Step 2: Generate the second AI span}

Continue the following hybrid text from the exact point where it stops.

Requirements:
\begin{lacpdpromptitems}
    \item Write only the continuation.
    \item Do not repeat any part of the prefix.
    \item The prefix ends in the middle of a sentence.
    \item Continue the unfinished sentence naturally.
    \item Do not restart the sentence.
    \item Continue the same topic, style, tone, and language
    as the prefix.
    \item Write \texttt{\{ai\_segment\_2\_min\_words\}} to
    \texttt{\{ai\_segment\_2\_max\_words\}} words.
    Aim for approximately
    \texttt{\{ai\_segment\_2\_target\_words\}} words.
    \item The word-count requirement is important.
\end{lacpdpromptitems}

Prefix:\par
\texttt{\{human\_part\_1\}}\allowbreak%
\texttt{\{ai\_segment\_1\}}\allowbreak%
\texttt{\{human\_part\_2\}}

Continuation:
\end{lacpdpromptbox}

\paragraph{Construction constraints.}
Boundary positions are stratified into three target ranges during
construction, as shown in Table~\ref{tab:app-boundary-strata}.
These ranges specify relative positions within the source window
and target sampling proportions.

\begin{table}[tbp]
\centering
\small
\caption{Target boundary-position strata during corpus construction.}
\label{tab:app-boundary-strata}
\begin{tabular}{lll}
\toprule
Stratum & Relative position & Target proportion \\
\midrule
Early  & 25\%--40\% & Approximately one third \\
Middle & 40\%--60\% & Approximately one third \\
Late   & 60\%--75\% & Approximately one third \\
\bottomrule
\end{tabular}
\end{table}

The generation configuration uses a temperature of $0.7$ and
top-$p$ of $0.9$. We retain generation status, prompt versions,
and token-audit information. During prefix construction, human
prefixes are required to contain at least two complete sentences.
H0 texts must contain at least 128 tokens, and the target AI spans
must contain at least 40 tokens. For cuts within a sentence, at
least eight valid words are retained on each side of the cut.
Where a right human context is required, it contains at least
64 tokens. B1 and B2 use source windows of 512 or 1,024 tokens,
whereas B3-2 and B3-3 use 1,024-token windows. Candidate boundaries
are constrained to avoid splitting words, named entities,
numbers, dates, or abbreviations.

\paragraph{Corpus size and data splits.}
The cleaned GPT-5.5 candidate pool contains 68,997 records,
with the condition distribution shown in
Table~\ref{tab:app-candidate-pool}. Before creating the final
splits, we remove two B1 records whose AI spans have zero length.
We group the remaining 68,995 records by
\texttt{source\_family\_id} and partition them into training,
validation, and test sets with target proportions of $7{:}1{:}2$.
This grouping keeps each source document and its derived examples
in the same split. Table~\ref{tab:app-corpus-splits} gives the
resulting statistics.

\begin{table}[tbp]
\centering
\small
\caption{Condition counts in the cleaned candidate pool, before
removing the two B1 records with zero-length AI spans.}
\label{tab:app-candidate-pool}
\begin{tabular}{rrrrrr}
\toprule
H0 & B1 & B2 & B3-2 & B3-3 & Total \\
\midrule
12,000 & 20,000 & 19,999 & 10,998 & 6,000 & 68,997 \\
\bottomrule
\end{tabular}
\end{table}

\begin{table}[tbp]
\centering
\small
\caption{Statistics of the complete GPT-5.5 corpus after removing
the two invalid records and grouping examples by source family.}
\label{tab:app-corpus-splits}
\begin{tabular}{lrrrrrr}
\toprule
Split & Total & H0 & B1 & B2 & B3-2 & B3-3 \\
\midrule
Training
& 48,297 & 8,434 & 14,073 & 14,018 & 7,639 & 4,133 \\
Validation
& 6,900 & 1,198 & 1,924 & 1,996 & 1,169 & 613 \\
Test
& 13,798 & 2,368 & 4,001 & 3,985 & 2,190 & 1,254 \\
\midrule
Total
& 68,995 & 12,000 & 19,998 & 19,999 & 10,998 & 6,000 \\
\bottomrule
\end{tabular}
\end{table}

\paragraph{Evaluation subsets and detector adaptation.}
The experimental evaluations use the following fixed,
condition-balanced subsets, sampled without replacement within
each condition:
\begin{itemize}[leftmargin=*,itemsep=2pt,topsep=4pt]
    \item \texttt{test\_bal200}:
    200 documents per condition, totaling 1,000 documents.
    This fixed subset is used for all reported GPT-5.5 evaluation
    results, including the main comparison, ablation study, and
    hyperparameter sensitivity analysis.
\end{itemize}

The default configuration is selected on the validation split
and fixed before evaluation on this held-out test subset.
The hyperparameter sensitivity analysis is conducted post hoc
and is not used to revise the default configuration.

Using fixed subsets gives equal representation to the five
conditions and controls the computational cost of
sentence-by-sentence Gemma-3-1B scoring and subsequent CPU-based
dynamic programming. The remaining test documents are retained
in the full test resource. Continued adaptation of the local
detector uses 4,000 human--LLM sentence pairs.

\paragraph{Claude-generated datasets.}
To evaluate transfer across text-generation models, we use
Claude-generated continuations based on SQuAD, WP, and XSum.
These datasets contain 98, 100, and 100 documents, respectively,
covering question answering, stories, and news. They primarily
represent a single authorship transition. Our evaluation uses
their preprocessed sentence lists and source labels.

\paragraph{Public datasets.}
We further evaluate generalization using RoFT and Hybrid Essays:
\begin{itemize}[leftmargin=*,itemsep=5pt,topsep=4pt]
    \item \textbf{RoFT.}
    Introduced in \emph{RoFT: A Tool for Evaluating Human
    Detection of Machine-Generated Text}, this dataset was
    designed to study whether humans can identify the transition
    from human writing to machine generation. Each example
    combines a human-written prefix with a language-model
    continuation, producing a single authorship transition.
    The data cover New York Times news articles, Reddit
    WritingPrompts stories, Recipe1M+ recipes, and U.S.
    presidential speeches. Continuations are generated by models
    including GPT-2, GPT-2 XL, fine-tuned GPT-2 XL, and CTRL,
    with GPT-3 Davinci also used in some experiments. The full
    collection contains 7,257 machine-generated continuations
    and 21,646 filtered human boundary annotations. We evaluate
    on a subset of 5,000 unique documents.

    \item \textbf{Hybrid Essays.}
    This dataset is associated with \emph{TriBERT: Towards
    Automatic Boundary Detection for Human-AI Collaborative
    Hybrid Essay in Education}. Its human-written texts come
    from the Automated Student Assessment Prize (ASAP-AES)
    corpus, which contains essays written by U.S. students in
    grades 7--10 in response to eight writing prompts. The
    dataset creators retain essays longer than 100 words and
    remove examples containing anonymized entities. They
    randomly remove sentences from the original essays and ask
    ChatGPT to fill the missing content using the surrounding
    context. The resulting patterns include
    $\mathrm{H}\rightarrow\mathrm{L}$,
    $\mathrm{L}\rightarrow\mathrm{H}$,
    $\mathrm{H}\rightarrow\mathrm{L}\rightarrow\mathrm{H}$,
    $\mathrm{L}\rightarrow\mathrm{H}\rightarrow\mathrm{L}$,
    $\mathrm{H}\rightarrow\mathrm{L}\rightarrow
      \mathrm{H}\rightarrow\mathrm{L}$,
    and
    $\mathrm{L}\rightarrow\mathrm{H}\rightarrow
      \mathrm{L}\rightarrow\mathrm{H}$,
    covering one to three authorship transitions. The full
    dataset contains 17,136 hybrid essays: 7,488 with one
    boundary, 6,429 with two boundaries, and 3,219 with three
    boundaries. Each essay contains approximately 287.6 words
    and 13.7 sentences on average, and approximately 65.3\% of
    the sentences are AI-generated. We use the official test
    set of 2,560 essays to evaluate sentence-level authorship
    prediction and boundary detection under multiple transitions.
\end{itemize}


\subsection{Baselines}
\label{app:baselines}

\paragraph{Sentence-level and heuristic baselines.}
\textbf{SenPred} applies a sentence-level source-prediction rule
independently to each sentence
~\citep{kushnareva2024roft}. A change between adjacent predicted labels
is treated as an authorship boundary, and consecutive LLM-labeled
sentences form predicted AI fragments.

\textbf{Voting} aggregates available sentence-level source predictions
using majority voting
~\citep{zhang2024machine}. We derive authorship boundaries from changes
in the resulting binary label sequence and form contiguous LLM-authored
spans from consecutive LLM labels.

\textbf{TextTiling} detects the boundaries of lexical-cohesion changes between neighboring text blocks~\citep{hearst1997texttiling}. We apply TextTiling to the ordered sentence sequence and assign each predicted segment a source label by applying the deterministic two-group thresholding to the set of segment-level means computed from the shared frozen-detector scores; the group with the larger mean is labeled LLM and the other Human. We assign the resulting labels to the sentences, derive authorship boundaries from label changes, and merge adjacent LLM-labeled segments into maximal LLM-authored spans without using ground-truth labels.

\paragraph{Score-based partial-text detection.}
\textbf{PaLD-scores} is a score-based adaptation inspired by
\emph{Detection of Text Partially Written by Large Language Models}
~\citep{lei2025pald}. It uses PaLD-derived scores to identify
LLM-written portions and delineates source labels and contiguous AI
spans from score variations. We use the name PaLD-scores to distinguish
this adaptation from the complete PaLD pipeline.

\paragraph{Change-point detection.}
We compare three change-point approaches for partitioning the
sentence-score sequence ~\cite{li2026segmenting}:

\begin{itemize}[leftmargin=*,itemsep=4pt,topsep=4pt]
    \item \textbf{VCP} uses an unweighted squared-error segment cost,
    so all sentences contribute equally to segment fitting.

    \item \textbf{WCP} uses a sentence-length-weighted squared-error
    segment cost and a complexity penalty to select the number of
    change points. The weight of each sentence is proportional to the
    square of its token count.

    \item \textbf{WCP+AIC} combines the length-weighted segment cost of
    WCP with an AIC-style criterion for selecting the number of change
    points. Its complexity multiplier is fixed at $r=2.0$, following
    the baseline implementation.
\end{itemize}

\paragraph{LA-CPD settings.}
In the unknown-count setting, LA-CPD uses dynamic programming to
optimize cut locations for each candidate cut count. It then evaluates
an AIC-style criterion using the original segment residuals, excluding
the boundary reward.

We also report \textbf{LA-CPD oracle-$K$} as a diagnostic setting. It
fixes the number of statistical cuts to $K=K^\star$, where
$K^\star$ denotes the reference number of final authorship boundaries.
The model still predicts the cut locations and source labels; it does
not receive the true statistical cut set or source labels. Because
adjacent statistical segments may receive the same label and be merged,
the statistical cut count and the final authorship-boundary count need not coincide.




\subsection{Experimental Settings and Evaluation Metrics}
\label{app:parameter-settings}

\paragraph{Data Input and Evaluation Protocol.}
The evaluation program receives preprocessed sentence lists and their
corresponding source labels. The main experiment reads
\texttt{sampled\_sentence} and \texttt{source\_label} and maps human and LLM
labels to 0 and 1, respectively. In the path that recomputes detector scores,
documents containing at most one sentence are skipped.

Under the unknown-count setting, the statistical partition is selected using
the sentence-score sequence and the inference parameters. The known-count
setting is evaluated separately as the LA-CPD oracle-$K$ comparison, where the
reference number of authorship boundaries is provided while cut locations and
source labels remain predicted by the model.

The reported metrics operate on different outputs. Sentence accuracy,
WindowDiff, and AI-fragment evaluation use the delineated sentence-level
authorship labels. Cut-count error compares the number of reference
authorship boundaries with the number of statistical cuts:
\begin{equation}
\mathrm{CE}
=
|C|-|\widehat{B}|,
\qquad
\mathrm{AbsCE}
=
\left|
|C|-|\widehat{B}|
\right|.
\label{eq:app-count-error}
\end{equation}

A positive $\mathrm{CE}$ indicates fewer statistical cuts than reference
boundaries, whereas a negative value indicates more statistical cuts. For an
individual document, $\mathrm{CE}=0$ means that the predicted and reference
boundary counts are equal. Since positive and negative errors can cancel when
averaged across documents, $\mathrm{AbsCE}$ is also reported to measure the
magnitude of count errors.

The final authorship boundary set $\widehat{C}$ is delineated from changes in
the predicted sentence labels. It can be smaller than $\widehat{B}$ when
adjacent statistical segments receive the same authorship label.

\paragraph{Local-Detector Adaptation.}
The local detector uses Gemma-3-1B as its backbone, with LoRA
adapters applied to \texttt{q\_proj}, \texttt{k\_proj},
\texttt{v\_proj}, and \texttt{o\_proj}. The main experiment
continues adaptation from an existing LoRA checkpoint using
4,000 human--LLM sentence pairs. The training configuration
is summarized in Table~\ref{tab:detector-training-config}.

\paragraph{Sentence-Level Accuracy (Acc).}
For a document containing $n$ sentences, accuracy is
\begin{equation}
\mathrm{Acc}
=
\frac{1}{n}
\sum_{i=1}^{n}
\mathbf{1}[\widehat{y}_i=y_i],
\label{eq:app-c-accuracy}
\end{equation}
where $y_i$ and $\widehat{y}_i$ are the reference and predicted
source labels of sentence $i$. We compute accuracy for each
document and report the mean across documents. Higher values
indicate better sentence-level authorship prediction.

\paragraph{WindowDiff (WD).}
WindowDiff compares the numbers of predicted and reference
authorship boundaries within sliding windows along the sentence
sequence. Boundaries are obtained from changes in the corresponding
sentence labels. Lower WD indicates closer agreement between the
predicted and reference segmentations.

\paragraph{AI-Fragment F1 (AI-F1).}
AI-F1 evaluates how well contiguous AI spans are detected and
localized. We extract maximal contiguous runs of AI-labeled
sentences from both reference and predicted source labels.
For a predicted fragment $P$ and a reference fragment $G$,
their intersection over union is
\begin{equation}
\operatorname{IoU}(P,G)
=
\frac{|P\cap G|}{|P\cup G|},
\label{eq:app-c-fragment-iou}
\end{equation}
where interval lengths are measured in sentences.

Within each document, predicted fragments are processed in
descending order of their mean sentence score. Each prediction
is greedily matched to the unmatched reference fragment with
the highest IoU. A match is accepted only when
$\operatorname{IoU}>0.5$, and each fragment can participate in
at most one match. Accepted matches count as true positives
(TP), unmatched predictions as false positives (FP), and
unmatched reference fragments as false negatives (FN).

We sum TP, FP, and FN across the evaluation set and compute
micro-averaged precision, recall, and F1:
\begin{equation}
\mathrm{Precision}
=
\frac{\mathrm{TP}}{\mathrm{TP}+\mathrm{FP}},
\qquad
\mathrm{Recall}
=
\frac{\mathrm{TP}}{\mathrm{TP}+\mathrm{FN}},
\label{eq:app-c-fragment-pr}
\end{equation}
\begin{equation}
\mathrm{AI\text{-}F1}
=
\frac{2\mathrm{TP}}
{2\mathrm{TP}+\mathrm{FP}+\mathrm{FN}}.
\label{eq:app-c-fragment-f1}
\end{equation}

AI-F1 ranges from zero to one, with higher values indicating
better fragment detection. On human-only documents, incorrectly
predicted AI fragments contribute to FP, whereas correctly
predicting no fragment does not increase TP. If both the
reference and predicted fragment sets are empty across the
entire evaluation set, precision, recall, and AI-F1 are defined
as one. If exactly one of these sets is empty, all three
metrics are defined as zero.

\subsection{Complete Ablation Results}
\label{app:full-ablation}

Table~\ref{tab:full-ablation-results} reports the complete
component ablation results, including WCP+AIC as a reference.
LA-CPD (Full) denotes LA-CPD+AIC.
The variant without $E(c;h)$ sets $\alpha=0$;
the variant without length weighting uses uniform weights;
and the variant without AIC replaces AIC-based selection
with a MAD complexity penalty.

\begin{table*}[t]
\centering
\small
\setlength{\tabcolsep}{5pt}
\renewcommand{\arraystretch}{1.1}
\caption{Complete component ablation results on the held-out
GPT-5.5 test subset of 1,000 documents.
Best and second-best results among the reported methods
are shown in \textbf{bold} and \underline{underlined},
respectively.}
\label{tab:full-ablation-results}

\begin{tabular}{lcccc}
\toprule
Variant
& Acc $\uparrow$
& WD $\downarrow$
& AI-F1 $\uparrow$
& B3-2 F1 $\uparrow$ \\
\midrule
LA-CPD (Full)
& \textbf{0.796} & \underline{0.231}
& \underline{0.600} & \textbf{0.338} \\
LA-CPD w/o $E(c;h)$
& 0.715 & 0.279 & 0.519 & 0.241 \\
LA-CPD w/o Length Weighting
& 0.781 & \textbf{0.223} & 0.516 & 0.185 \\
LA-CPD w/o AIC
& \underline{0.789} & 0.250
& \textbf{0.605} & \underline{0.325} \\
\midrule
WCP+AIC (reference)
& 0.747 & 0.270 & 0.552 & 0.280 \\
\bottomrule
\end{tabular}
\end{table*}

\subsection{Complete Cross-Distribution Generalization Results}
\label{app:generalization}

\begin{figure*}[t]
\centering
\includegraphics[width=\textwidth]
{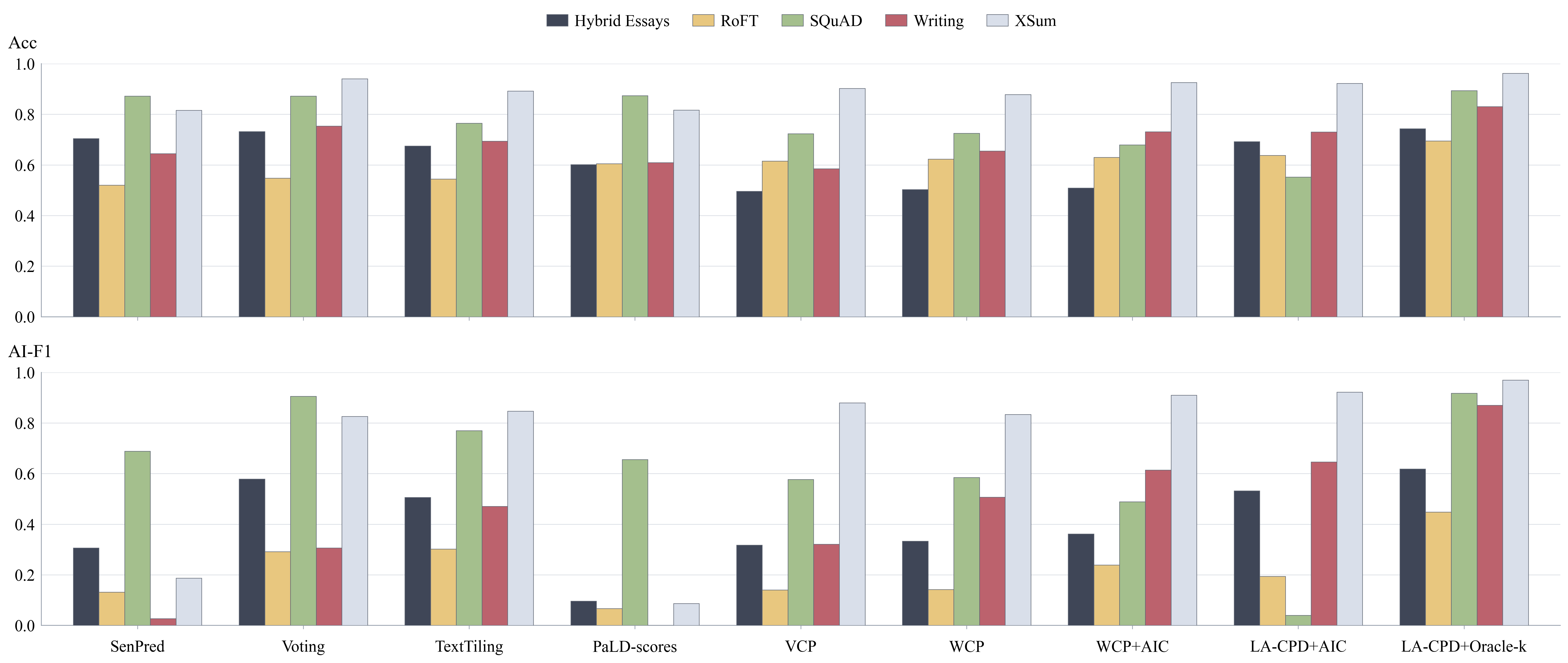}
\caption{Complete cross-distribution generalization results on
Hybrid Essays, RoFT, Claude--SQuAD, Claude--WritingPrompts, and
Claude--XSum. The top and bottom panels report sentence-level accuracy
and AI-fragment F1, respectively. Bars are grouped by method, and colors
identify the evaluation datasets. Missing results are left blank.
Higher Acc and AI-F1 indicate better performance. LA-CPD oracle-$K$
uses the ground-truth boundary count and is included only as a
diagnostic setting.}
\label{fig:generalization-all-metrics}
\end{figure*}

Figure~\ref{fig:generalization-all-metrics} reports the complete
cross-distribution generalization results in terms of sentence-level
accuracy and AI-fragment F1. The corresponding datasets and evaluation
protocols are described in Section~\ref{sec:experiments}. WindowDiff
(WD) is additionally reported for Hybrid Essays and RoFT, but is
unavailable for the Claude-based evaluations.

On Hybrid Essays, LA-CPD+AIC achieves an Acc of 0.692, an AI-F1
of 0.532, and a WD of 0.310. Its AI-F1 is higher than those of
VCP, WCP, and WCP+AIC, whereas Voting obtains the highest AI-F1
among the non-oracle methods. On RoFT, LA-CPD+AIC achieves the
highest Acc among methods with unknown boundary counts, reaching
0.638. Its WD of 0.210 is close to that of WCP, while its AI-F1
remains lower than those of Voting and TextTiling.

The Claude-based evaluations show larger performance differences
across domains. On Claude--WritingPrompts, LA-CPD+AIC achieves an
AI-F1 of 0.646, exceeding the reported non-oracle baselines, although
its Acc of 0.730 is lower than that of Voting. On Claude--XSum,
LA-CPD+AIC obtains an AI-F1 of 0.922, the highest result among
methods with unknown boundary counts, while its Acc of 0.922 is
lower than the 0.940 obtained by Voting. In contrast, LA-CPD+AIC
performs poorly on Claude--SQuAD, with an Acc of 0.552 and an
AI-F1 of 0.040. These results show that the generalization advantage
of LA-CPD+AIC is not uniform across generation models and text
domains.

Providing LA-CPD with the ground-truth boundary count improves both
Acc and AI-F1 on all five datasets. Because the cut locations and
source labels remain predicted in this setting, the comparison
isolates the effect of boundary-count selection. For example,
LA-CPD oracle-$K$ improves Acc and AI-F1 on Claude--SQuAD from
0.552 and 0.040 to 0.894 and 0.918, respectively. It also improves
AI-F1 from 0.194 to 0.448 on RoFT and from 0.646 to 0.870 on
Claude--WritingPrompts. These improvements indicate that
boundary-count selection is an important source of cross-distribution
error, although generalization also depends on detector-score quality,
text domain, generation model, and document characteristics.

\subsection{Supplementary Analysis of the Regime-Return Discount}
\label{app:regime-return}

We examine the optional regime-return discount under a
MAD-based complexity penalty on the held-out GPT-5.5 test set
of 1,000 documents. The parameter $\rho$ controls the penalty
discount for the second and subsequent statistical cuts.
Setting $\rho=1$ applies no discount, whereas $\rho<1$
reduces these penalties.

We vary $\rho\in\{1.00,0.75,0.50,0.35\}$ while fixing
$h=3$, $\alpha=0.5$, $\beta=1$, and the MAD penalty
coefficient $c=2$. This configuration differs from the
AIC-based configuration used in the main experiments.
The results are reported in Table~\ref{tab:rho-ablation}.

\begin{table}[t]
\centering
\small
\setlength{\tabcolsep}{5pt}
\renewcommand{\arraystretch}{1.1}
\caption{Sensitivity to the regime-return discount $\rho$
under the MAD-based complexity penalty.
Other parameters are fixed to
$h=3$, $\alpha=0.5$, $\beta=1$, and $c=2$.}
\label{tab:rho-ablation}
\begin{tabular}{lcccc}
\toprule
Metric & $\rho=1.00$ & $\rho=0.75$
       & $\rho=0.50$ & $\rho=0.35$ \\
\midrule
Acc $\uparrow$
& 0.756 & 0.747 & 0.737 & 0.731 \\
H0 Acc $\uparrow$
& 0.852 & 0.840 & 0.829 & 0.810 \\
AI-F1 $\uparrow$
& 0.519 & 0.512 & 0.510 & 0.513 \\
B3-2 AI-F1 $\uparrow$
& 0.237 & 0.249 & 0.264 & 0.276 \\
\bottomrule
\end{tabular}
\end{table}

Reducing $\rho$ from 1.00 to 0.35 increases B3-2 AI-F1
from 0.237 to 0.276, but decreases overall Acc from
0.756 to 0.731 and H0 Acc from 0.852 to 0.810.
Overall AI-F1 does not improve, changing from 0.519
to 0.513. These results indicate that stronger discounts
benefit AI-fragment delineation in the B3-2 condition
at the expense of overall and human-only sentence accuracy
under the evaluated MAD configuration.
They do not establish a corresponding benefit under
AIC-based selection; the main experiments use $\rho=1$.

\end{document}